\documentclass{article}

\usepackage[preprint]{neurips_2024}

\usepackage[utf8]{inputenc} 
\usepackage[T1]{fontenc}    
\usepackage{hyperref}       
\usepackage{url}            
\usepackage{booktabs}       
\usepackage{amsfonts}       
\usepackage{nicefrac}       
\usepackage{microtype}      
\usepackage{xcolor}         
\usepackage{graphicx}
\usepackage{subfig}
\usepackage{hyperref}
\usepackage{amsmath}

\usepackage{tabularx}
\usepackage{comment}
\usepackage{listings}
\usepackage{stfloats}
\usepackage{makecell}

\definecolor{codegreen}{rgb}{0,0.6,0}
\definecolor{codegray}{rgb}{0.5,0.5,0.5}
\definecolor{codepurple}{rgb}{0.58,0,0.82}
\definecolor{backcolour}{rgb}{0.95,0.95,0.92}

\lstdefinestyle{mystyle}{
  backgroundcolor=\color{backcolour}, commentstyle=\color{codegreen},
  keywordstyle=\color{magenta},
  numberstyle=\tiny\color{codegray},
  stringstyle=\color{codepurple},
  basicstyle=\ttfamily\footnotesize ,
  breakatwhitespace=false,         
  breaklines=true,                 
  captionpos=b,                    
  keepspaces=true,                                  
  numbersep=5pt,                  
  showspaces=false,                
  showstringspaces=false,
  showtabs=false,                  
  tabsize=2
}
\newcommand*{\affmark}[1][*]{\textsuperscript{$#1$}}

\title{SMAC-R1: The Emergence of Intelligence in Decision-Making Tasks}

\author{
Yue Deng$^{1}$,
Weiyu Ma$^2$,
Yuxin Fan$^3$,
Ruyi Song$^1$
Yin Zhang$^1$,
Haifeng Zhang$^2$,
Jian Zhao\\
\affmark[1]College of Computer Science and Technology, Zhejiang University \\
\affmark[2]Institute of Automation, Chinese Academy of Sciences \\
\affmark[3]School of Mathematics and Statistics, Xi'an Jiaotong University \\
\texttt{devindeng@zju.edu.cn, maweiyu2022@ia.ac.cn, fanyuxin@xjtu.stu.edu.cn} \\
\texttt{3220105868@zju.edu.cn, zhangyin98@zju.edu.cn, haifeng.zhang@ia.ac.cn} \\
}
\begin{document}

\maketitle

\begin{abstract}

StarCraft Multi-Agent Challenge (SMAC) has been one of the most commonly used experimental environments in multi-agent reinforcement learning (MARL), where the specific task is to control a set number of allied units to defeat enemy forces. 
Traditional MARL algorithms often require interacting with the environment for millions of steps to train a parametric model, of which the resulting policies are typically non-interpretable with weak transferability.
In this paper, we introduce SMAC-R1 which is based on the Qwen2.5-7B-Base LLM distilled from DeepSeek-Coder-v2.5-236B.
Similar to online reinforcement learning after behavior cloning in offline learning process, in our pipeline, agents leverage the DeepSeek LLM to generate decision tree code by providing task descriptions, and the agents are further self-reflected using feedback from the rewards provided by the environment.
Based on that, we augment the generated scripts to fine-tune a small LLM, Qwen2.5-7B-Base, to distill the decision-making ability via Supervised Fine-Tuning (SFT) and enhance the script generation ability by the Group Relative Policy Optimization (GRPO) algorithm.
We conduct experiments in the original 23 SMAC tasks and 10 newly-designed tasks to demonstrate that our method can produce high-quality, interpretable decision trees with minimal environmental exploration.
Moreover, these scripts exhibit strong transferability, successfully applying to homogeneous SMAC environments without modification.
We believe this approach offers a new direction for solving decision-making tasks and domain-specific LLM training pipelines in the future.
\end{abstract}

\section{Introduction}

The StarCraft Multi-Agent Challenge (SMAC)\citep{smac} has emerged as a widely adopted benchmark for the evaluation of multi-agent reinforcement learning (MARL) algorithms.
Built upon the real-time strategy game StarCraft II, SMAC provides a highly challenging environment in which each agent controls a distinct unit on the battlefield.
Agents must operate in partially observable and dynamic environments where effective coordination is crucial to achieving common objectives.
With access only to local information, agents are required to collaborate while contending with adversarial units governed by built-in AI opponents.
The environment offers a diverse set of tasks, ranging from small-scale skirmishes to large-scale, complex battles, making it an ideal testbed for assessing agent cooperation, coordination, and competitive behavior in MARL settings.
Consequently, SMAC has become a standard benchmark for evaluating the performance of cutting-edge algorithms, including VDN\citep{vdn}, QMIX\citep{qmix}, QPLEX\citep{qplex}, and MAPPO\citep{mappo}, which have shown their effectiveness in this domain.
    
Despite the impressive performance of MARL in solving SMAC tasks, it has several notable drawbacks and challenges:
\begin{itemize}
	\item
	\textbf{Low exploration efficiency}: MARL requires extensive interaction with the environment to learn an optimal policy, which often demands significant training time and computational resources. In complex environments, an agent may need to perform millions of interactions to acquire an effective strategy.
    \item 
    \textbf{Lack of model interpretability}: Since deep learning relies on complex neural networks, MARL models are often considered black boxes, making it difficult to explain the decision-making process. In certain applications, such as autonomous driving or healthcare, this lack of interpretability can raise concerns about safety or regulatory compliance. In contrast, humans may benefit from white-box reasoning such as human-in-the-loop and bad case correction scenarios when the decision-making process is completely exposed to humans.
    \item
    \textbf{Poor policy transferability}: MARL models are typically trained in specific environments, and they struggle to generalize or be directly applied to different environments. The transferability of policies is weak, requiring retraining or significant adjustments to adapt to new tasks or environments.
\end{itemize}

Traditional rule-based decision tree models can indeed overcome the aforementioned issues in the white-box method, but building such models requires a significant amount of human expertise and prior knowledge, which leads to high labor costs. 
For a long time, there has been no effective method to integrate the strengths of both approaches.
    
With the advent of Large Language Models (LLMs)\citep{instructgpt,gpt4_report,llama,bai2023qwen,glm}, especially code-oriented LLMs\citep{deepseek-coder,qwen-coder}, the ability of models to generate code has become increasingly robust, providing a novel approach for their integration.
During pre-training, LLMs learn from vast amounts of human experience and knowledge, particularly in the domain of coding, laying a strong foundation for models to generate code that can solve complex tasks.
Agents can take task descriptions as input, generate corresponding decision tree code, and iteratively improve the decision tree based on task feedback.
This paradigm not only addresses the challenge of lacking a reward function during LLM training but also overcomes the issue of the non-interpretability of policies generated by reinforcement learning.

Meanwhile, in recent advancements in reinforcement learning from human feedback (RLHF), frameworks such as LLAMA Factory \citep{zheng2024llamafactory}, OpenRLHF \citep{hu2024openrlhf}, and VERL \citep{sheng2024hybridflow} have emerged as pivotal tools for aligning AI systems with human preferences. OpenRLHF provides an open-source platform for implementing RLHF, enabling researchers to fine-tune models using human feedback in a scalable and efficient manner. Similarly, the VERL framework focuses on verifiable and interpretable RLHF, ensuring that the learned policies are not only effective but also transparent and robust. Complementing these frameworks, we distill the problem-solving ability from the DeepSeek-Coder-V2.5-236B to small Qwen2.5-Coder-7B, thereby reducing computational costs while maintaining performance. Together, these approaches represent significant strides in creating AI systems that are both aligned with human values and practical for real-world deployment.
    
Taking the SMAC task, we propose our SMAC-R1 model based on Qwen2.5-7B-base, through the SFT and GRPO enhancement pipeline after the plan-code-critic process.
We are excited to report that SMAC-R1, with minimal environmental exploration, produces a high-quality white-box decision tree model on the majority of the original 23 SMAC maps and 10 newly-designed maps.
Moreover, this model exhibits strong transferability, as it can be applied to similar few-shot SMAC environments without any modifications.
Through visualization, we observe that the agents provided by our SMAC-R1 model acquire the micro-skills required for SMAC, such as kite and focus fire. 
Moreover, these agents prioritize macro tactics and are less prone to exploiting specific loopholes in the game environment to achieve victory, resulting in higher strategy robustness.
Additionally, focusing on the SMAC coding tasks, the GRPO results shows completely opposite results compared to DeepSeek models that the response body containing strategy and skills tend to be short yet precise for SMAC instead of long answers for math-solving problems.
We believe this presents a new perspective for addressing decision-making tasks.

\section{Related Work}

\subsection{StarCraft II Game AI} 
Research in artificial intelligence for StarCraft II has advanced significantly since DeepMind introduced the PySC2 learning environment in 2017 \citep{pysc2}. This platform, together with Blizzard's release of anonymized game replays, provided a standardized interface for developing and testing AI agents. A landmark achievement occurred in 2019 with the introduction of AlphaStar \citep{alphastarnature}, which reached Grandmaster status and defeated elite human competitors, showcasing the capabilities of reinforcement learning in intricate settings.
    
Following the AlphaStar milestone, explorations diverged into multiple innovative pathways. Mini-AlphaStar \citep{liu2021introduction} was developed to streamline input variables while preserving performance. Concurrent studies, such as TG \citep{liu2021efficient} and HierNet-SC2 \citep{liu2022onefficient}, investigated more efficient reinforcement learning strategies. AlphaStar Unplugged \citep{starcraft2unplugged} marked a substantial progression in offline reinforcement learning utilizing human game replays. Further developments were made with TStarBotsX \citep{Tstarbot-x} and SCC \citep{scc}, enhancing federated learning approaches; notably, SCC achieved a significant victory against the Chinese champion "Time" in 2020.
    
Recent breakthroughs also include DI-star, facilitating the deployment of StarCraft II AI on home computers, and ROA-Star \citep{ROA-Star}, which augmented AlphaStar's training framework with goal-conditioned exploiters and sophisticated opponent modeling techniques, defeating the world champion "hero". The introduction of LLM Play SC2\citep{llm_play_sc2} represents the pioneering use of large language models for macro-strategic decision-making in StarCraft II, thereby opening new horizons for LLMs applications in real-time strategy games.
    
\subsection{LLM Agents in Complex Game Environments}
The advent of advanced language models like GPT-3.5 \citep{instructgpt} has significantly propelled research on LLM agents forward, particularly in complex game environments. This research spans a wide range of game types, from classic board games to open-world video games, showcasing the versatility of LLM agents.
    
In board games, ChessGPT\citep{chessgpt} demonstrated LLM agents' ability to understand and play strategic games. The social deduction game Werewolf presented unique challenges in strategic communication and decision-making, with recent studies \citep{werewolf_kim, werewolf_xu} exploring LLM agents' capabilities in this complex, multi-agent environment.
    
Video game applications have been particularly diverse and impressive. The MineDojo environment \citep{minedojo} facilitated projects like GITM \citep{ghostintheminecraft} and Voyager \citep{voyager} in Minecraft, demonstrating LLM agents' ability to navigate and perform tasks in complex 3D environments. PokéLLMon \citep{pokemon} showcased their effectiveness in turn-based tactical games, emphasizing the importance of external knowledge retrieval and iterative policy refinement.
    
The Cradle framework \citep{cradle} introduced a novel approach allowing LLM agents to interact with various software and games through a unified interface of screenshots and keyboard/mouse inputs. This demonstrated the potential for general-purpose game-playing agents across multiple commercial video games and software applications. CivRealm \citep{civrealm} presented a complex, imperfect-information general-sum game environment, challenging both the learning and reasoning capabilities of AI agents.

\subsection{Code as Action in LLM Agents}
Recent research has explored the use of executable code as a unified action space for LLM agents, demonstrating significant potential in enhancing agent capabilities across various domains. This approach, often referred to as "code as action" or "code as policies," has shown promising results in several key areas.
    
The CodeAct framework \citep{executable_code_actions} proposed using executable Python code to consolidate LLM agents' actions. This approach outperformed traditional JSON or text-based action spaces, offering greater flexibility and the ability to compose multiple tools. The study demonstrated up to a 20 \% higher success rate compared to widely used alternatives.
    
In the realm of robotics, Chen et al. \citep{code_as_policy} introduced "code as policies" for robot control. In this approach, LLMs generate Python code to process perception outputs and parameterize control primitives. This method enabled robots to exhibit spatial-geometric reasoning, generalize to new instructions, and prescribe precise values based on context, demonstrating effectiveness across multiple robot platforms.
    
The Eureka algorithm \citep{eureka} leveraged LLMs for human-level reward design in reinforcement learning tasks. By performing evolutionary optimization over reward code, Eureka generated reward functions that outperformed expert human-engineered rewards across a diverse suite of RL environments, including complex tasks like dexterous pen spinning.
    
The success of these "code as action" approaches suggests a promising direction for developing more capable and adaptable LLM agents, with potential applications in robotics, game AI, and other complex decision-making domains.

\section{Method}

\begin{figure}[t]
    \centering
    \includegraphics[width=0.95\linewidth]{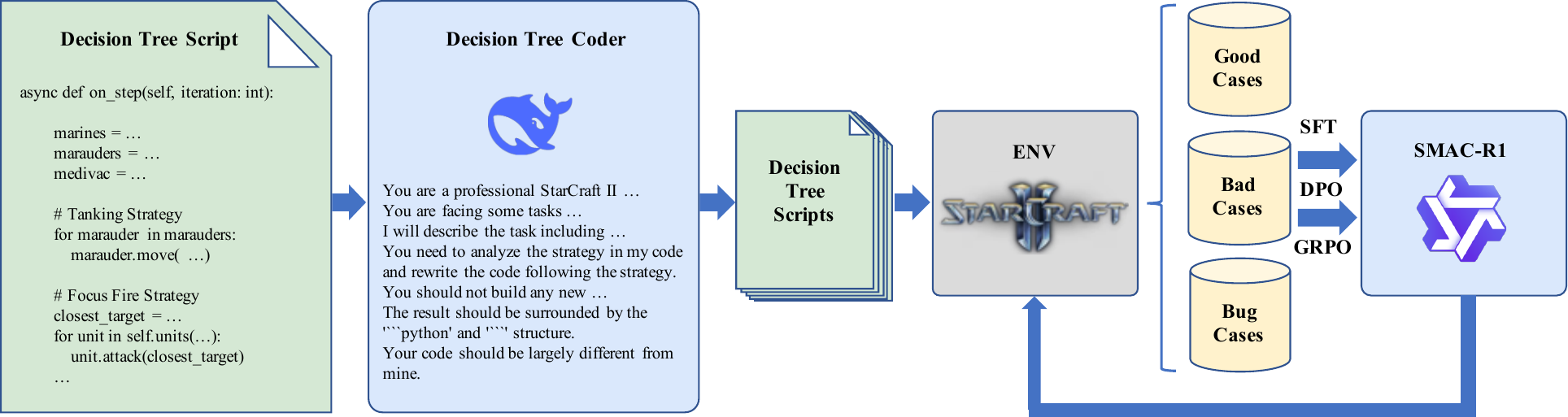}
    \caption{The overall architecture of distilling, fine-tuning, and improving process of the student LLM (SMAC-R1). The generated decision-tree scripts are augmented by a coder LLM and generate a bunch of scripts with similar strategies. The scripts are divided into good, bad, and bug cases by evaluation on SMAC. Then the SMAC-R1 is fine-tuned by SFT, DPO, and finally GRPO by the data and iteratively fine-tuned by the scripts generated by our SMAC-R1 model.}
    \label{fig:GRPO}

\end{figure}

This section presents the training pipeline of our proposed SMAC-R1 model designed to address SMAC tasks, including generating decision tree codes from coder LLMs, distilling decision-making ability to smaller student LLM, using generated scripts, and enhance reasoning capabilities via the GRPO algorithm. By harnessing the reasoning, planning, and code-generation capabilities of the DeepSeek LLM, our framework effectively solves the majority of SMAC micro-management tasks. Furthermore, we transfer the policy-generation ability to our SMAC-R1 model, which is based on the Qwen2.5-7B-Base, through SFT and the GRPO algorithm. The training pipeline, depicted in Figure \ref{fig:pcc} and Figure \ref{fig:GRPO}, encompasses script generation, script augmentation, SFT training, and GRPO-based improvement. The subsequent sections will provide a detailed explanation of these components.

 \subsection{Architecture}
    
\paragraph{Overview} LLMs have acquired extensive knowledge about StarCraft II and learned python-sc2 and other SC2 scripts during their pretraining phase. Building upon this foundation, our framework addresses the code generation task for the python-sc2 package\footnote{\url{https://github.com/BurnySc2/python-sc2}}, which provides a comprehensive wrapper for StarCraft II gameplay. Unlike traditional MARL models on the pysc2 platform, the scripts from the python-sc2 package are open-loop control scripts that require the LLM to plan and anticipate game scenarios before execution. The extensive workload of planning, code generation, bug identification, and bug fixing places significant strain on a single LLM. To address this, in the decision-tree generation phase, our framework divides the process into three critical modules: the strategy planning model, the decision-tree code generation model, and the summarization critic model. Then the successfully generated scripts that can win the SMAC task are augmented to increase the diversity and used to SFT and DPO a student LLM. The SFT and DPO processes provide student LLMs standard architecture of response script, which should prevent from API mis-invocation and async function definition. Then the fine-tuned LLMs are further improved by GRPO by the actions sampled from the student itself to increase the winning rate of generated scripts.

\begin{figure}[htb]
    \centering
    \includegraphics[width=0.99\linewidth]{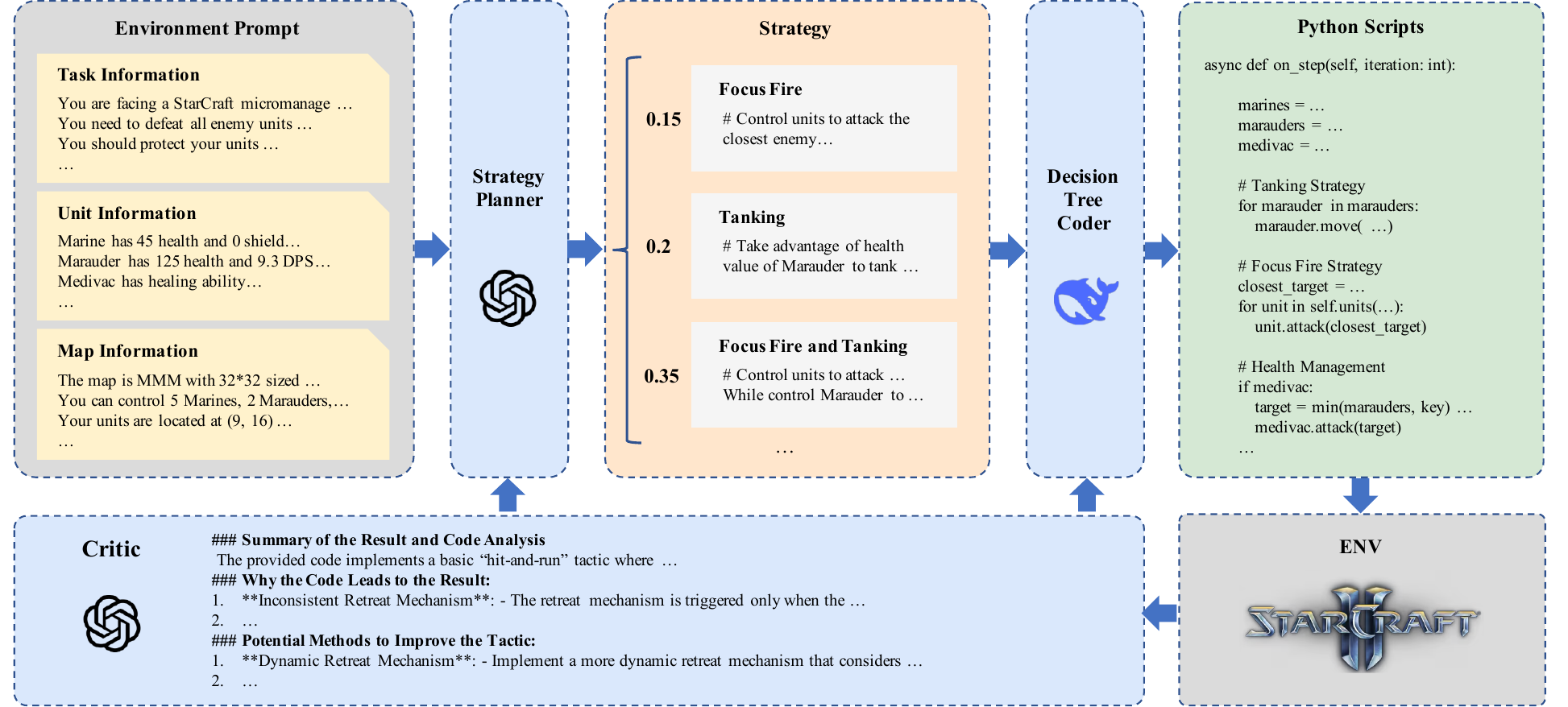}
    \caption{The overall architecture of our proposed LLM-SMAC framework. The framework takes the information of units, maps, and tasks as the environment prompts. Then, the planner generates a strategy conditioned on the environment, and the coder generates Python scripts following the strategy. The scripts are tested on the SMAC environment, and the results with the strategy and the coder are fed into the critic to analyze the refinement suggestion. Finally, the planner and the coder generate new strategies and codes in closed loops.}
        \label{fig:pcc}
\end{figure}

\paragraph{Decision Tree Script Generation} As illustrated in Figure \ref{fig:pcc}, to solve a SMAC task, unit and map information are analyzed and presented as an environment prompt for the planner LLM. Additionally, the current learned skill set and historical strategy traces are provided to the planning LLM to help generate a strategy skeleton. The planner then suggests skill options, describes each skill, and defines the conditions for using them. This decision-tree representation serves as the input for the LLM coder, which generates the corresponding Python scripts. Due to LLM hallucinations and the existence of multiple ways to implement a strategy skeleton, the coder iteratively generates and tests the code on SMAC maps. Based on the correctness and performance of the code, the SMAC simulation process provides stack traces of bugs or key results, such as win rates, scores, and damage statistics. These results and the scripts serve as the input to the critic module to analyze the reasons for high or low performance or identify specific bugs. Then, the critic provides suggestions for refining the strategy, as well as code corrections when bugs are present. The critic LLM also offers final recommendations on whether to revise the planned strategy or improve the current implementation, guiding the workflow for the next round, either to the planner or the coder.

\paragraph{SFT and GRPO} During the distillation and fine-tuning phase, we utilize synthetically generated Python scripts alongside a coder LLM to produce novel decision tree-based scripts. A secondary coder LLM is employed to strategically rewrite these scripts, preserving their core tactical logic while altering their implementation details. The revised scripts are then evaluated in SMAC environments through rollout simulations, where they are categorized into three classes: good cases (high win rate), bad cases (low win rate), and bug cases (containing bugs identified via stack traces). To mitigate data imbalance, duplicate scripts are eliminated by comparing their abstract syntax tree (AST) representations. Our SMAC-R1 model is subsequently trained in two stages: first through supervised fine-tuning (SFT) on the successful scripts, then via Direct Preference Optimization (DPO) using both successful and suboptimal scripts to minimize API misuse risks. Then, the Group Relative Policy Optimization (GRPO) algorithm refines the model’s code generation capability, aligning it with task-specific performance metrics in SMAC scenarios.

\subsection{Planner-Coder-Critic}

\paragraph{Planner}The planner provides the coder with an overall strategy and detailed descriptions of the corresponding skills. Due to the nature of open-loop control, the planner can only offer general, coarse-grained strategy suggestions without detailed information about the agent units or the maps in each iteration. Therefore, the planner must refine these suggestions based on unit and map information. In our framework, unit details are pre-scraped from Liquipedia, where information about unit characteristics and usage can be automatically retrieved and incorporated into the environment prompt. Meanwhile, map information, including size, available operational areas, and terrain features, is also summarized and added to the prompt.
    
To leverage the in-context learning capability of LLMs, our framework stores historical skills and their corresponding rollout win rates as nodes and return values in an MCTS (Monte Carlo Tree Search) structure. The planner must not only plan based on the environmental prompt but also consider past planning decisions by either selecting a new strategy or adding a skill to an existing historical strategy. 

\begin{equation}
	\begin{small}
    ST = (s_0,..., s_n) = LLM_{planner}(I_{u} + I_{m} + h(s_i, r_i)),
    \end{small} 
\end{equation}

where the output strategy $ST$, composed of a sequence of skills $s_i$, is generated by the LLM planner, which takes into account unit information $I_{u}$, map information $I_{m}$, and historical data. The historical data includes each skill $s_i$ in the current skill pool and their associated score $r_i$.  
    
\paragraph{Coder}Given the strategy generated by the planner, the coder module is tasked with producing the corresponding Python code using the python-sc2 package for testing. However, the performance of the generated code depends on several factors: the strategy provided, the interpretability of the LLM itself, and the quality of the training data that used for training the LLM. To enhance the strategy’s performance and facilitate the translation of the strategy into code, the planner must not only provide the skills but also specify the conditions for invoking each skill through detailed prompts: 

\begin{equation}
    C = LLM_{coder}(ST, LLM_{critic}(C', G)), 
\end{equation} 
where the code $C$ is generated based on the strategy $ST$ and improvement suggestions from the critic LLM, which are derived from the previously generated code $C'$ and the simulation results $G$.
    
The generated code is then evaluated using StarCraft II simulations. The average win rate and additional metrics such as scores, damage dealt, damage taken on health, and damage absorbed by shields are calculated over 10 simulation runs. If any exceptions occur during testing, the stack traces of these exceptions are provided to the critic LLM for analysis, and the win rate is set to 0. 
\begin{equation} 
    G = SMAC(C, map)
\end{equation}
    
\paragraph{Critic}When exceptions occur in the generated Python code, the critic model is tasked with identifying the bugs and potential fixes, reducing the burden on the coder model. The critic analyzes the generated code $C$ along with the traceback information to pinpoint the cause of the exception. Due to LLM hallucinations and API changes in the python-sc2 package, the coder model may mistakenly invoke deprecated or incorrect APIs from older versions of python-sc2. While increasing model parameters and using a more capable coder LLM can significantly reduce the likelihood of such errors, the code still needs to be checked against the latest official python-sc2 API documentation.
    
If no exceptions occur during the rollout, the test results are provided to the critic model, which analyzes the reasons behind the current performance, identifying both the factors contributing to high performance and lessons from any underperformance. Ultimately, the critic provides explanations, potential improvements, and recommendations on whether to adjust the strategy or refine the current approach ($LLM_{critic}(C', G)$) to guide the next steps in the pipeline.
    
After re-entering the planning phase, the simulation results are backpropagated and incorporated into the planner’s historical data by updating the average score for each skill: 
\begin{equation} 
    r_i = {(r_i \times \Gamma_i + G)}/{(\Gamma_i + 1)} 
\end{equation} 
where $r_i$ is the score of skill $s_i$, and $\Gamma_i$ represents the number of times skill $s_i$ has been used historically.

\subsection{SMAC-R1 training}

According to the planner-coder-critic framework description in the previous section, successful Python script generation depends on the success of the three modules. Consequently, the failure of any module will result in the final script generation. Meanwhile, the generating difficulty also results in the insufficient amount of script file collection for fine-tuning. Therefore, we utilize another LLM to expand the generated scripts and tune our SMAC-R1 in single-step generation format instead of multi-step reasoning.

\paragraph{Script data augmentation} We use the Python script generated by the coder to fine-tune our SMAC-R1 via data generation, duplicated file removal, SFT based on good cases, DPO based on good cases and bad cases and GRPO according to the SMAC reward function. To do so, we first use another LLM to iteratively imitate and rewrite the provided code, which provides a bunch of new scripts. After that, the newly generated Python scripts are tested via the SMAC environment and classified into good cases (winning rate larger than 80\%), bad cases (winning rate less than 80\%), and bug cases.

\paragraph{Duplicated file removal} Because the generated codes derive from very few successful examples, these files have a certain probability of being duplicated. We design an algorithm to evaluate and minimize redundancy within code datasets by leveraging structural normalization and similarity computation. We first do the code normalization that parses source code into its Abstract Syntax Tree (AST) representation using the Python AST module. A customized AST transformer, CodeNormalizer, standardizes variable names, function names, and class names. This transformation ensures that semantically identical but syntactically varied code fragments are normalized to a common structure. Then, the normalized ASTs are converted to string representations to calculate the similarity ratio between different code fragments by structural comparison. By comparing structural representations, the algorithm outputs a similarity percentage to judge the similarity between two script files in order to control the diversity of datasets. This approach ensures that dataset redundancy is analyzed at a structural level, providing a robust method for identifying semantically equivalent code with minimal syntactic influence. 

\paragraph{Training Pipeline} After removing the redundancy script code, we firstly fine-tune the SMAC-R1 base model (Qwen2.5-7B-Base) by the good cases via OpenRLHF \citep{hu2024openrlhf} package. To mitigate the problem of API misuse, the signatures, the usage, and the effects of the functions described in the official documentation are crawled, formatted and added to the training dataset. Apart from that, we also crawl the python-sc2 scripts from the aiarena where the StarCraft community players upload their mandatory edited scripts, and provide the codes to LLM to learn other API invocation. After the SFT process, we continue to fine-tune the reference model by DPO algorithm according to the equation:
\begin{equation}
 	\mathcal{L}_{DPO}=-\mathbb{E}_{(x,y_c,y_r)\sim \mathcal{D}}\left[\log\sigma\left(\beta\log\frac{\pi_\theta(y_c|x)}{\pi_{ref}(y_c|x)} -\beta\log\frac{\pi_\theta(y_r|x)}{\pi_{ref}(y_r|x)}\right)\right],
\end{equation} 
where $\mathcal{D}$ is the dataset, $\sigma$ is the sigmoid function, $\beta$ is a hyper-parameter, $y_c$ is the chosen response, $y_r$ is the rejected response, and $x$ is the input prompt. 

After the DPO fine-tuning, we leverage the GRPO \citep{shao2024deepseekmath} algorithm to improve our SMAC-R1 model via reinforcement learning. According to the GRPO algorithm formation, the system prompts are treated as the question $q$, and the scripts generated are the answers $o_i$. Then, GRPO samples a group of outputs $\{o_1, o_2, ..., o_G\}$ from the old policy $\pi_{\theta_{old}}$ in each iteration. Then our SMAC-R1, $\pi_\theta$, is optimized by maximizing the following objective:
\begin{equation}
\begin{split}
	\mathcal{J}_{GRPO}=&\mathbb{E}_{\{o_i\}_{i=1}^G\sim\pi_{\theta_{old}}(O|q)}\\
	 &\frac{1}{G}\sum_{i=1}^G(\min(\frac{\pi_\theta(o_i|q)}{\pi_{\theta_{old}}(o_i|q)}A_i, clip(\frac{\pi_\theta(o_i|q)}{\pi_{\theta_{old}}(o_i|q)},1-\epsilon, 1+\epsilon)A_i)-\beta\mathbb{D}_{KL}(\pi_\theta||\pi_{ref}))
\end{split}
\end{equation}
where the $\epsilon$ and $\beta$ are hyper-parameters, and $A_i$ is the advantage that is computed using a group of rewards $\{r_1,r_2,...,r_G\}$ corresponding to the outputs within each group:
\begin{equation}
	A_i=\frac{r_i-mean(\{r_1,r_2,...r_G\})}{std{\{r_1,r_2,...r_G\}}}
\end{equation}
The SMAC environment is a naturally suitable reward provider, so we  define the score of the GRPO algorithm as follows:
\begin{enumerate}
	\item $-1$: The valid Python script cannot be retrieved from the response content of SMAC-R1.
	\item $-0.5$: There are bugs in the script, and the test process fails to terminate normally.
	\item $0\sim1$: The winning rate of the generated code in SMAC tasks.
\end{enumerate}

\section{Experiment}

In this section, we evaluate our proposed framework on 23 scenarios from the SMAC micro-management tasks, which require the use of various individual and combined skills. We also test our SMAC-R1 on 10 newly-designed similar scenarios. The results demonstrate that through the planning, coding, and critic phases, the LLMs are able to select appropriate skills, generate Python code that complies with python-sc2 requirements, and successfully win most the combat scenarios. Meanwhile, the fine-tuned SMAC-R1 is able to infer correct scripts and successfully make decisions in the few-shot scenarios.
    
We validate the generated code within the StarCraft II client using 10 different random seeds. Both the times taken for coding are recorded as part of the results.
For the sake of convenience, we use the open-source DeepSeek-V2.5 (deepSeek-Coder-v2.5-236B) as the base model for the planner, coder, and critic modules in API mode. However, these models can be replaced with other advanced alternatives, such as GPT-4.
    
\subsection{Main Results}

We present the results of win rates, coding rounds, coding rounds after SFT and DPO, and the final strategies planned by the LLM for each map in Table \ref{table:main_result}. As shown in the table, the LLM-generated code achieves near-perfect win rates on easier tasks and competitive results on more difficult tasks. Leveraging the LLM's prior knowledge, the planner is able to select optimal strategies with relative ease, and thanks to the high coding capabilities of the DeepSeek model, most correct Python scripts are generated within 30 rounds of iteration.
    
\paragraph{Strategy Planning} The planner LLM outlines the strategy framework by combining skills and defining the conditions under which they should be used. According to the results, "Focus Fire" emerges as one of the most common and effective skills across SMAC maps, as it quickly reduces the number of enemy units. For many of the easier tasks, once the planner identifies and selects the "Focus Fire" skill, the likelihood of generating a winning code script significantly increases. Conversely, on maps where focus fire is less effective, such as the corridor map, the LLM requires more rounds to produce alternative skills after receiving feedback from the rollout results.
    
\begin{figure}[!ht]
    \centering
    \includegraphics[width=0.99\linewidth]{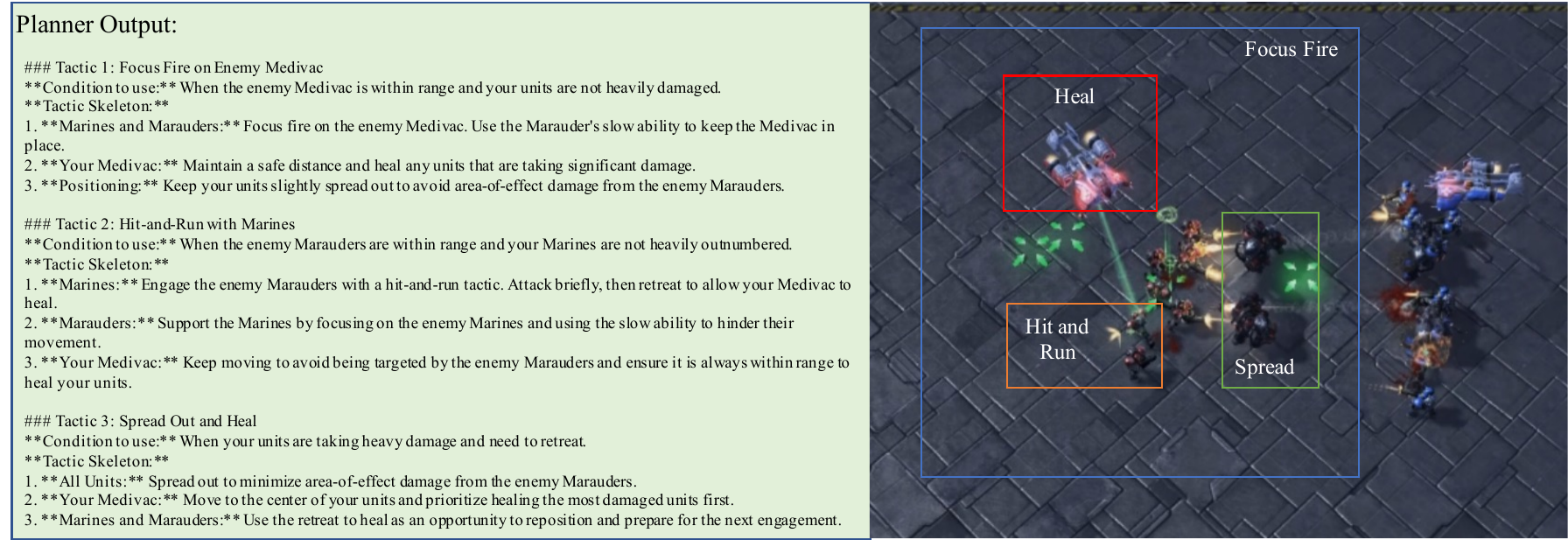}
    \caption{An example planner output that provides three potential skills: focus fire, hit and run, and Spread out and heal for the medivac unit. The planner also describes how to use each skill and when to use the skill. }
    \label{fig:planner1}
\end{figure}

As illustrated in Figure \ref{fig:planner1}, certain skills can conflict with one another. For example, the focus fire skill directs all units to attack the weakest enemy, while the kiting skill commands units to retreat to the backline. At the same time, the positioning skill might push some units forward to the front. As a result, a unit may receive conflicting commands, such as one attack order and two movement orders, which leads to inconsistent actions during the rollout. By refining the conditions for each skill, the planner makes the overall strategy more precise and provides the coder with a well-structured decision tree for further implementation.
    
\begin{figure}[h!]
    \centering
    \includegraphics[width=0.99\linewidth]{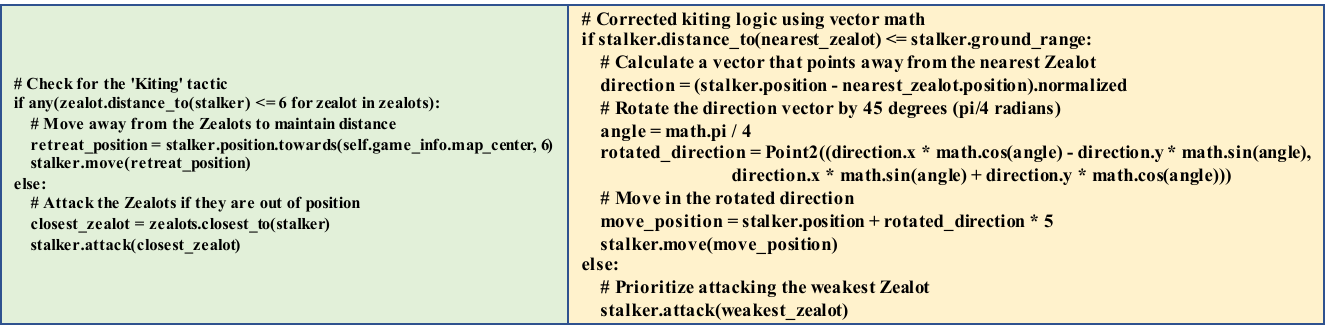}
    \caption{An example coder output from the first round and after the critic promotion round given a main strategy skeleton. Both of the two example codes follow the strategy, but the latter code is more precise.}
    \label{fig:coder2}
\end{figure}

\paragraph{Code Generation} Given the strategy planned by the planner and the refinement suggestions from the critic model, the coder module generates Python code in accordance with the python-sc2 package. Due to the coder’s powerful yet still limited coding ability and the diversity of possible implementation methods, multiple rounds of code generation are required to fix any bugs in the decision tree script and to improve performance based on the rollout results.

Starting from a strategy skeleton, we present the coding result from the first round and the refined code following the promotion prompt from the critic module. Except for value adjustments and skill additions, the coder can enhance the implementation by accounting for more situational factors. As illustrated in Figure \ref{fig:coder2}, in the kiting strategy, agents should retreat in the opposite direction to maintain a minimum distance from the closest enemy. However, on smaller maps, units may get stuck at the map’s edge, resulting in the retreating behavior being taken but failing. After several rounds of in-context learning improvements, the critic identifies that the planned strategy is sound, but the implementation lacks detail. Consequently, the coder begins dynamically calculating the retreat direction in the kiting method, improving the strategy's overall effectiveness.

\paragraph{SMAC-R1 SFT and DPO} We fine-tune the Qwen2.5-7B-Base model by the scripts generated by the DeepSeek-coder-v2.5-236B model and their augmented scripts. We evaluate our model on the SMAC tasks and calculate the average coding times among 5 successful scripts which result in a higher than 90\% winning rate. According to Table \ref{table:main_result}, the coding round with the critic module of DeepSeek is less than 30, but after the SFT and DPO rounds, the student model can generate correct codes in less than 5 rounds. Apart from that, we evaluate the student model on 10 newly designed but with similar features. According to Table \ref{fig:few-shot}, the student model can lock on the terrain advantage in the 3rp\_vs\_4zl map, the shield and healing management in 1mr1md\_vs\_1sc map, focus fire in 3mr\_3mr and 5hd\_vs\_5hd scenarios and Split in 6mr\_vs\_1ac scenarios. This phenomenon also states the few-shot reasoning ability of the student LLM after the SFT and DPO.
    
\begin{table}[!h]
    \centering
    \begin{small}
    \begin{tabular}{cccccl}
    \hline
        Map Name & WR \% & \makecell[c]{DeepSeek\\-236B} & \makecell[c]{Qwen2.5-7B\\(SFT+DPO)} & Planned Strategy \\ \hline
        3m & 100\% & 2 & 1.4 & Focus Fire \\
        8m & 100\% & 5 & 1 & Focus Fire \\
        5m\_vs\_6m & 0\% & - & 1.4 & Initial Forming, Focus Fire \\
        8m\_vs\_9m & 90\% & 22 & 2 & Initial Forming, Focus Fire \\
        10m\_vs\_11m & 100\% & 19 & 1 & Focus Fire, Grouping \\
        25m & 100\% & 32 & 1.2 & Initial Forming, Focus Fire \\
        27m\_vs\_30m & 90\% & 35 & 1.2 & Initial Forming, Focus Fire, Grouping \\
        2s3z & 100\% & 31 & 2 & Focus Fire, Grouping \\
        3s5z & 100\% & 22 & 3 & Focus Fire, Grouping \\
        1c3s5z & 100\% & 11 & 2.6 & Engagement, Hit and Run, Flanking \\
        3s5z\_vs\_3s6z & 0\% & - & - & Focus Fire, Grouping, Kiting \\
        2m\_vs\_1z & 100\% & 4 & 1 & Hit and Run \\
        3s\_vs\_3z & 100\% & 3 & 1 & Focus Fire, Hit and Run \\
        3s\_vs\_4z & 100\% & 14 & 1.2 & Focus Fire, Hit and Run \\
        3s\_vs\_5z & 100\% & 14 & 1.4 & Focus Fire, Hit and Run \\
        2c\_vs\_64zg & 100\% & 18 & 1.2 & Health and Shield Management, Terrain Advantage \\
        corridor & 100\% & 17 & 1 & Health and Shield Management, Terrain Advantage \\ 
        MMM & 100\% & 15 & 3.6 & Focus Fire, Hit and Run, Split out and Heal \\
        MMM2 & 0\% & - & - & Focus Fire, Hit and Run, Split out and Heal \\
        2s\_vs\_1sc & 100\% & 23 & 1.2 & Health and Shield Management, Terrain Advantage \\ 
        bane\_vs\_bane & 100\% & 14 & 2.4 & Split \\
        so\_many\_baneling & 100\% & 11 & 1 & Split \\
        6h\_vs\_8z & 0\% & - & - & Focus Fire, Grouping \\ \hline

    \end{tabular}
    \end{small}
    \caption{The winning rate over 10 times validation, the strategy planning rounds, the code generation rounds, and the planning results of each map in SMAC. We use the DeepSeek-Coder-V2.5 model with different parameters. The bugs are caused by the wrong API calls from the python-sc2 package in the generated codes. The results show the round number before the first successful generation for Deepseek and the average rounds of 5 successful generations for Qwen.}
    \label{table:main_result}
\end{table}

\begin{table}[!h]
    \centering
	\begin{small}
    \begin{tabular}{ccccl}
    \hline
        Agent Units & Enemy Units & Map & Critical Strategy \\ \hline
        1 Marauder and 1 Medivac & 1 Spine Crawler & Flat map & Health Management, Heal \\
        2 Void Ray  & 3 SporeCrawler & Flat map & Shield Management \\ 
        3 Marauder & 3 Marauder & Flat map & Focus Fire \\
        3 Reaper & 4 Zealot & Cliff & Terrain Advantage \\
        5 Hydralisk & 5 Hydralisk & Flat map & Initial Forming \\ 
        6 Marine & 1 Archon (Splash Damage) & Flat map & Split \\
        6 Marine & 1 SiegeTankSieged (Splash Damage) & Flat map & Split \\ 
        3 Stalker & 4 Zealot & Cliff & Kite, Path Finding \\
        3 Hellion & 24 Zergling & Flat map & Kite, Focus Fire \\
        3 Reaper & 24 Zergling & Flat map & Kite, Focus Fire \\
    \hline
    \end{tabular}
    \end{small}
    
    \caption{New maps to test few-shot abilities of student LLM. Each map is similar to some original SMAC maps in the aspect of terrain features, shield features, or strategies. }
    
    \label{fig:few-shot}
\end{table}

\paragraph{SMAC-R1 GRPO optimization}

\begin{figure}[!ht]
    \centering
    \includegraphics[width=\linewidth]{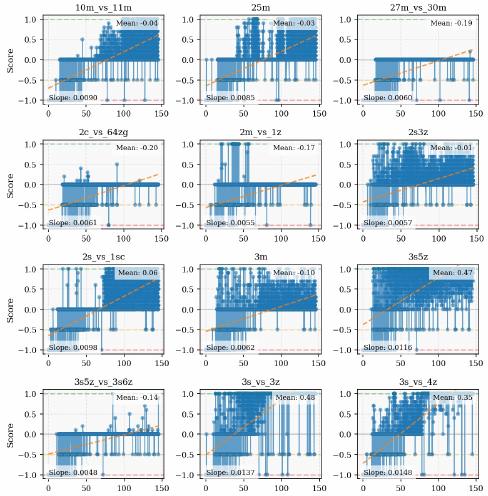}
    \caption{The learning curves among the 12/23 tasks in SMAC. The x-axis is the GRPO training steps and the y-axis is the average winning rate.(1/2)}
    \label{fig:grpo_results1}
\end{figure}

\begin{figure}[!ht]
    \centering
    \includegraphics[width=\linewidth]{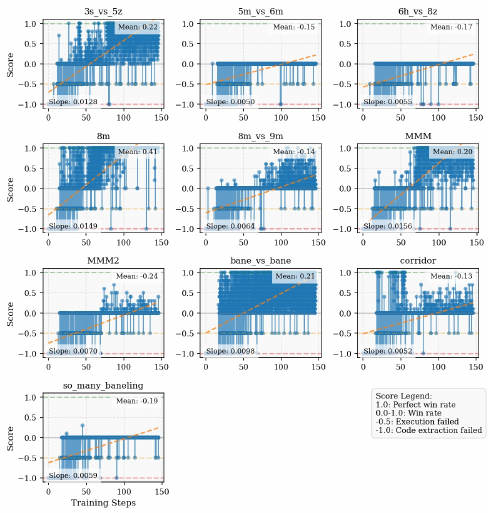}
    \caption{The learning curves among the 11/23 tasks in SMAC. The x-axis is the GRPO training steps and the y-axis is the average winning rate.(2/2)}
    \label{fig:grpo_results2}
\end{figure}

Apart from applying iterative DPO to our SMAC-R1 model, we also integrate the GRPO algorithm using the reward calculation described earlier. The reward curves for different SMAC maps are presented in Figure \ref{fig:grpo_results1} and Figure \ref{fig:grpo_results2}, while Table \ref{table:grpo_sta} provides the reward distribution statistics.

As shown in Figure \ref{fig:grpo_results1} and Figure \ref{fig:grpo_results2}, the learning curves in most scenarios begin at a score of -1, indicating that the generated Python code is either missing from the response or incorrectly formatted. After limited training steps, the curves quickly approach -0.5, suggesting that SMAC-R1 can produce scripts, albeit with errors. Subsequently, the curves rapidly reach 0 and continue rising toward 1. It is worth noting that on the map 3s5z\_vs\_3s6z and the MMM2 which are super difficult in task configurations, our SMAC-R1 fails to generate correct scripts with the DPO process, yet generates scripts with certain winning rates after the GRPO processes. In the tasks such as 5m\_vs\_6m, it is hard for SMAC-R1 to explore an optimal strategy during the training process, so we believe that the model will generate an increasing number of correct samples with additional training steps.

\begin{table}[!ht]
    \centering
    \begin{tabularx}{\textwidth}{>{\centering\arraybackslash}p{3cm}
    >{\centering\arraybackslash}X
	>{\centering\arraybackslash}X
	>{\centering\arraybackslash}X
    >{\centering\arraybackslash}X
    >{\centering\arraybackslash}X
    >{\centering\arraybackslash}X
    }
    \hline
        Map Name&  Count&   Mean&   std&  Min&  Max&  Median\\
      10m\_vs\_11m&   3995& -0.045& 0.483& -1.0&  1.0&     0.0\\
             25m&   3949& -0.028& 0.489& -1.0&  1.0&     0.0\\
      27m\_vs\_30m&   3953& -0.193& 0.367& -1.0&  0.2&     0.0\\
      2c\_vs\_64zg&   3968& -0.196& 0.375& -1.0&  0.5&     0.0\\
        2m\_vs\_1z&   3946& -0.174& 0.375& -1.0&  1.0&     0.0\\
            2s3z&   3994& -0.008& 0.432& -1.0&  1.0&     0.0\\
       2s\_vs\_1sc&   3987&  0.060& 0.517& -1.0&  1.0&     0.0\\
              3m&   4000& -0.096& 0.416& -1.0&  1.0&     0.0\\
            3s5z&   3984&  0.470& 0.681& -1.0&  1.0&     0.8\\
    3s5z\_vs\_3s6z&   3988& -0.144& 0.340& -1.0&  1.0&     0.0\\
        3s\_vs\_3z&   3969&  0.483& 0.726& -1.0&  1.0&     1.0\\
        3s\_vs\_4z&   3968&  0.349& 0.702& -1.0&  1.0&     0.5\\
        3s\_vs\_5z&   3961&  0.217& 0.636& -1.0&  1.0&     0.1\\
        5m\_vs\_6m&   3980& -0.150& 0.336& -1.0&  0.0&     0.0\\
        6h\_vs\_8z&   3969& -0.174& 0.355& -1.0&  0.0&     0.0\\
              8m&   3966&  0.406& 0.740& -1.0&  1.0&     1.0\\
        8m\_vs\_9m&   4008& -0.135& 0.393& -1.0&  1.0&     0.0\\
             MMM&   3953&  0.203& 0.732& -1.0&  1.0&     0.0\\
            MMM2&   3969& -0.243& 0.368& -1.0&  0.7&     0.0\\
    bane\_vs\_bane&   3962&  0.212& 0.626& -1.0&  1.0&     0.4\\
        corridor&   4012& -0.131& 0.391& -1.0&  1.0&     0.0\\
so\_many\_baneling&   4019& -0.188& 0.360& -1.0&  0.3&     0.0\\
\hline
    \end{tabularx}
    \caption{The number of testing instances and the GRPO training statistics, including the mean, std, min, max, and median value of each SMAC task.}
    \label{table:grpo_sta}
\end{table}

\section{Discussion}

\begin{figure}[!h]
    \centering
    \includegraphics[width=0.99\linewidth]{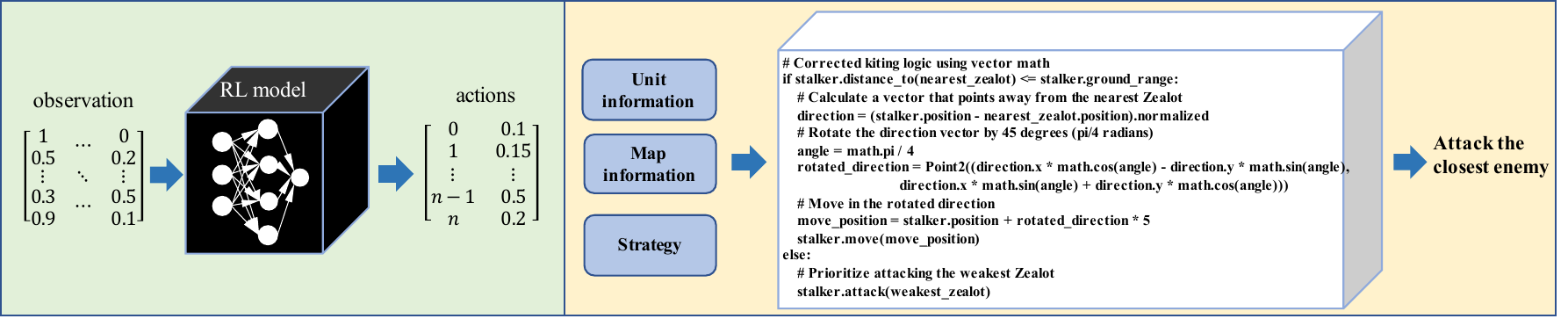}
    \caption{A model interpretability example comparison between the traditional reinforcement learning (black-box) and the decision tree script (white-box) decision-making processes.}
    \label{fig:explain}
    \vspace{-10px}
\end{figure}

\paragraph{White-box Decision-Making via LLM} Unlike traditional model-free reinforcement learning, which approximates state-action values, the decision-tree scripts generated by the coder explicitly represent skills and decision-making processes. As shown in Figure \ref{fig:explain}, utility networks from MARL can compute the maximum expected return for a specific action, but they fail to explain the relationship between actions taken and observations acquired. In contrast, by leveraging the LLM’s code generation and few-shot reasoning abilities, the planner generates the strategy skeleton in natural language, while the coder defines action logic in machine language with accompanying comments in natural language. This approach greatly enhances the interpretability of decision-making.

\begin{figure}[tb]
    \centering
    \subfloat[]{\includegraphics[width=0.32\columnwidth]{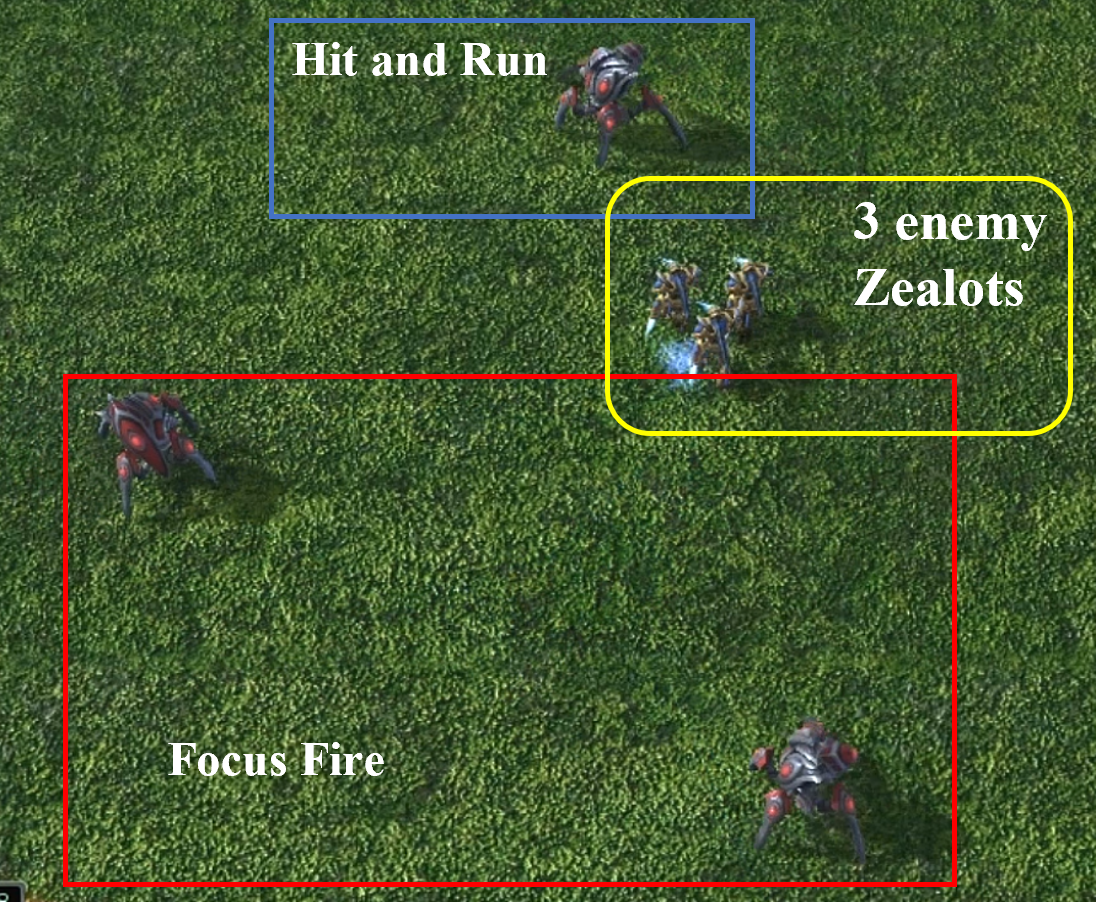}} 
    \hspace{2px}
    \subfloat[]{\includegraphics[width=0.32\columnwidth]{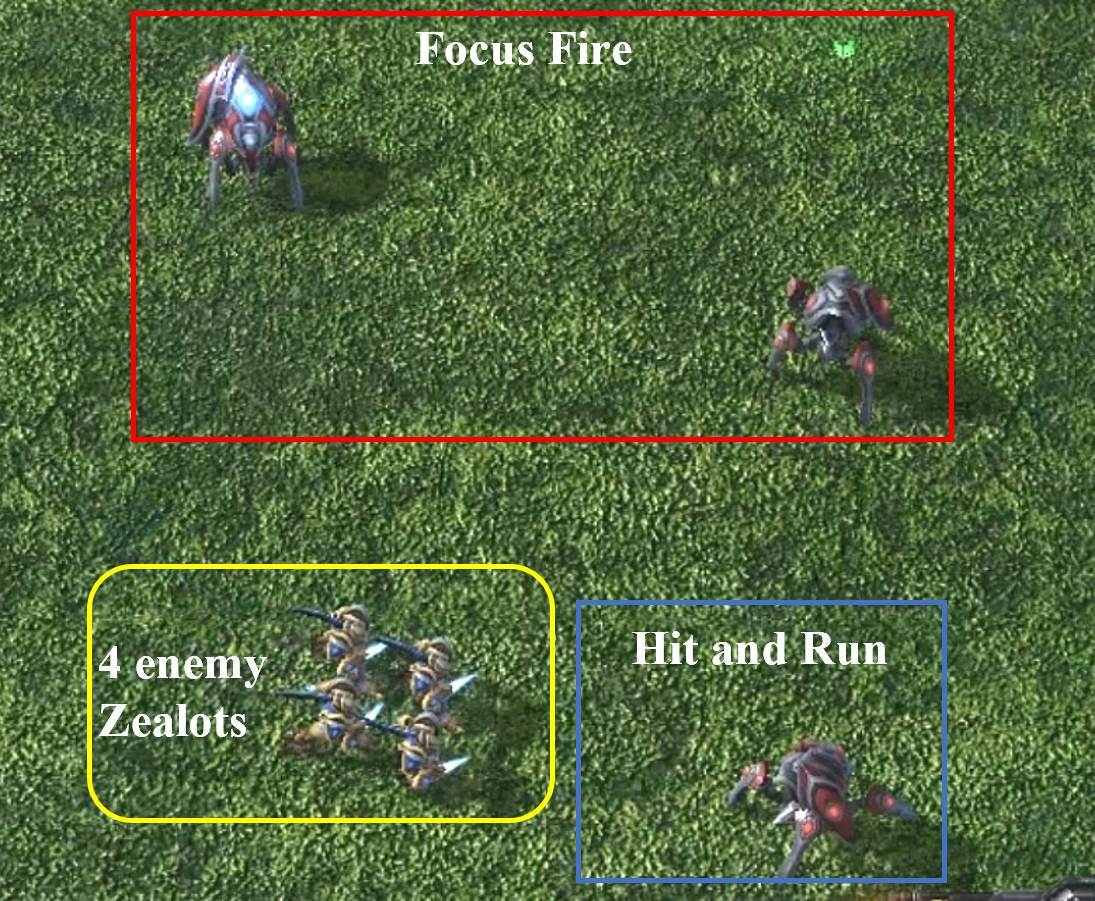}} 
    \hspace{2px}
    \subfloat[]{\includegraphics[width=0.32\columnwidth]{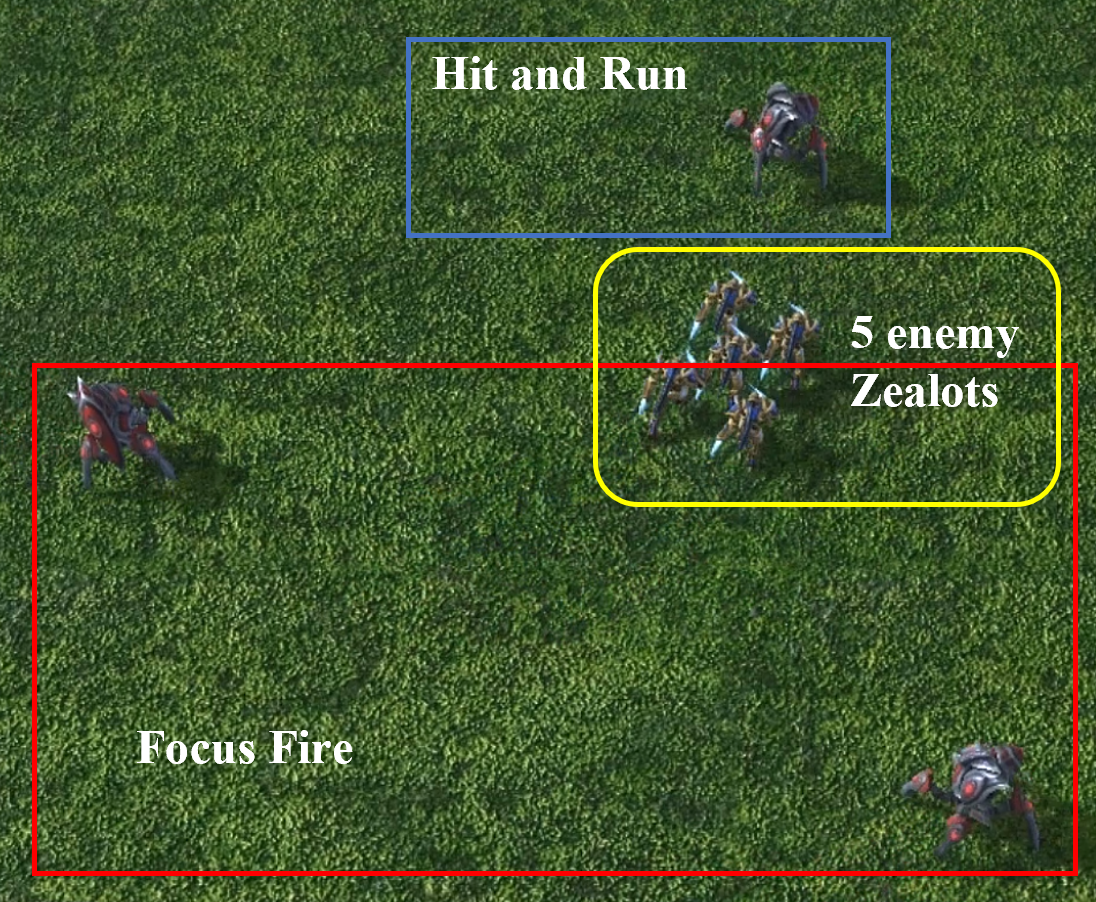}}
    \caption{An example indicates that the strategy learned from the 3s\_vs\_4z scenario is that one of the units is responsible for kiting, and another two units are responsible for attacking. Because the universal strategy is learned on the homogeneous scenario, the strategy is also suitable for 3, 4, and 5 enemy Zealots.}
    \label{fig:transfer} 
\end{figure}

\paragraph{Policy Transferability} The decision-tree script can be viewed as a set of time-series regulations triggered by different conditions. As a result, strategies designed for more difficult tasks can be directly transferred to similar tasks that feature analogous unit formations. For example, in video demonstrations, the strategy developed for the 8m scenario can be applied without modification to the 3m scenario. Similarly, as shown in Figure \ref{fig:transfer}, the policy generated for the 3s\_vs\_4z scenario (where three units form a triangle and one unit attracts enemy fire while the other two focus fire on the weakest target) can be directly reused in both the 3s\_vs\_3z and 3s\_vs\_5z scenarios. The strategy from 3s5z can also be effectively applied to the 2s3z scenario, offering a more precise solution than one learned directly from 2s3z.

\paragraph{Exploration Efficiency} Vanilla reinforcement learning models often suffer from low exploration efficiency, as they must train from scratch. Although some approaches in multi-agent reinforcement learning emphasize coordinated exploration techniques, these methods still explore at the action level, which limits their efficiency. In contrast, our SMAC-R1 framework leverages the prior knowledge of LLMs to explore at the skill level in the planner phase and at the decision-tree implementation level in the coder phase. This significantly enhances efficiency, allowing us to solve hard SMAC tasks in just tens of rounds of planning and code generation, whereas MARL algorithms typically require millions of timesteps over tens of thousands of episodes.

\begin{figure}[tb]
    \centering
    \subfloat[]{\includegraphics[width=0.49\columnwidth]{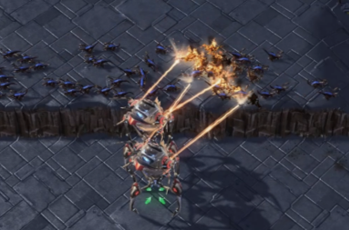}} 
    \hspace{2px}
    \subfloat[]{\includegraphics[width=0.49\columnwidth]{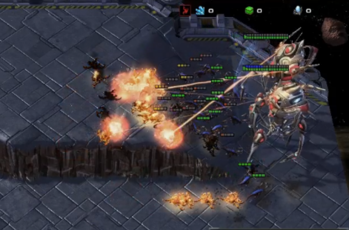}} 
    \caption{(a) The two Colossus units apply the cliff climbing skill to kite the enemy Zerglings given the map information. (b) The two Colossus move to the edge of the map to reduce the surrounding damage from the Zerglings.}
    \label{fig:map_info} 
\end{figure}

\paragraph{Map Information} Map descriptions provide crucial terrain advantages that enrich strategic planning. Some units, like the Colossus, can use the terrain to their advantage, such as climbing cliffs to kite enemies. Other units may use map edges to block choke points, reducing the risk of being surrounded by Zerglings. As illustrated in Figure \ref{fig:map_info}, two Colossi can exploit the terrain by climbing up and down a cliff while most enemy Zerglings remain on the other side. Without this map information, strategies might unintentionally rely on map edge advantages. Another example is the corridor scenario, where units should hold choke points to prevent being surrounded by Zerglings. Without the choke point, the generated code is hard to win the corridor combat.

\paragraph{Information on Units} While the training of LLMs already incorporates knowledge about units, explicitly providing comprehensive unit information for the specific map can enhance their effectiveness. The data crawled from Liquipedia includes essential attributes such as health, shield, and damage per second (DPS). This information enables the planner to perform statistical calculations, such as determining when to employ the kiting skill based on differing unit speeds. The retreat point is also influenced by the attacking range of units. Consequently, detailed unit information aids the planner module in generating more targeted strategies, allowing the coder to produce more precise decision-tree scripts.

\paragraph{Response Body Length}

\begin{figure}[!h]
    \centering
    \includegraphics[width=0.99\linewidth]{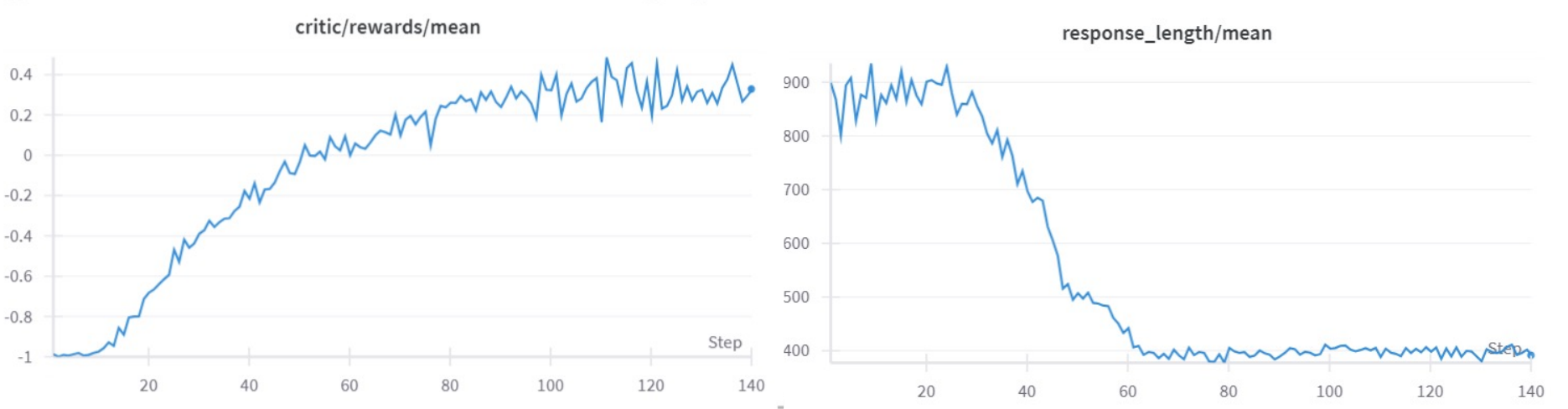}
    \caption{A comparison between the average winning rate and the length of response body during the GRPO training process.}
    \label{fig:length}
\end{figure}

As shown in Figure \ref{fig:length}, the average win rate exhibits a negative correlation with the length of the response body. The GRPO case study in the Appendix further supports this, indicating that longer preceding strategy analyses tend to produce more complex but less relevant implementations, ultimately leading to lower-quality code.

Unlike DeepSeek-R1, where the response body progressively lengthens, SMAC-R1 generates increasingly shorter responses. This contrast highlights a fundamental difference between SMAC strategy reasoning and mathematical problem-solving. In math problem-solving, agents need to explore solutions through extensive reasoning. However, in SMAC decision-making, unnecessary strategies lead to wasted actions and a lower win rate. For instance, in the 3m scenario, agents must consistently focus fire on a single opponent until it is eliminated. If one agent retreats prematurely, the overall firepower decreases, increasing the likelihood of losing. As a result, the final strategy converges to the simplest yet most effective approach, avoiding unnecessary complexity.

\section{Conclusion and Future Work}

This paper proposes SMAC-R1 to solve SMAC tasks through the planner-coder-critic pipeline and GRPO algorithm based on SMAC tasks. 
We firstly leverage LLM with more parameters to generate decision tree code and refine the decision tree based on feedback from the environment's rewards in the planner-coder-critic framework.
Our experiments demonstrate that LLM-SMAC can produce a high-quality white-box decision tree model with minimal environmental exploration in most SMAC tasks and strong policy transferability.
Based on that, we also fine-tune LLMs using generated scripts and apply GRPO reinforcement learning methods by the reward functions to further enhance their decision tree generation capabilities.
Results show that SMAC-R1 can significantly reduce the code generation rounds while maintaining the few-shot decision-making ability.
Inspired by the pipeline of decision tree generation, distillation, and enhancement via reinforcement learning, we believe that generating decision tree code through LLM should offer a new perspective for addressing decision-making tasks and see great potential for further exploration in this direction.

\clearpage
\bibliography{neurips_2024.bib}
\bibliographystyle{neurips_2024}

\clearpage
\appendix
\onecolumn

\section{Prompts}

In this section, we show the prompt in the planner, coder, and critic module. Because of the different code lengths and different map and unit descriptions, we use placeholders in upper case here.

\begin{table*}[h!]
\begin{small}
    \centering
    \begin{tabularx}{\textwidth}{|>{\centering\arraybackslash}X|}
    \hline
    \makecell[l]{    
    The map is 10m\_vs\_11m of 32*32 sized square map.\\
    The available area of the x-axis is from 0 to 32, and the y-axis is from 7 to 25.\\
    The enemy units are at (23, 16) point, and your units are at (9, 16) point initially.\\
    The enemy controls all the units to move and attack (9, 16) points along the way.\\
    There are no terrain advantages or choke points in this map.\\
    You can control 10 Marine units.\\
    The enemy controls 11 Marine units.\\
    }\\
    \hline
    \end{tabularx}
    \captionsetup{labelformat=empty}
    \caption{The map information description.}
    \end{small}
\end{table*}

\subsection{Planner}
\begin{table*}[h!]
\begin{small}
    \centering
    \begin{tabularx}{\textwidth}{|>{\centering\arraybackslash}X|}
    \hline
    \makecell[l]{    
    You are a StarCraft II player and a helpful assistant, and you are facing micro-management tasks.\\
I will describe the map and the units on the map. You should not build any new buildings or new units. \\
You may focus on micro-management tactics to win the combat, such as \\
positioning, initial formation, fire focusing, kiting, tanking, hit and run, health and shield management,\\
retreat, split, terrain advantages, or any combinations among them. \\
You should provide me the tactics in the format below:\\
\#\#\# Tactic 1: Tactic 1' name\\
**Condition to use:**\\
**Tactic Skeleton**\\
\\
\#\#\# Tactic 2: Tactic 2' name\\
**Condition to use:**\\
**Tactic Skeleton**\\
\\
Meanwhile, I will tell you history taken tactics, \\
you may add new tactics onto the history tactics by adding new tactic in the list or try a new tactic. \\
The task is TASK INFORMATION \\
You should provide me with at most 3 most important tactics and describe the chosen tactic skeleton \\
in detail according to the situations of your unit and enemy units.\\
You should also indicate the condition to use this tactic. \\
You should set some status variables for units to prevent skills conflict.\\
    }\\
    \hline
    \end{tabularx}
    \captionsetup{labelformat=empty}
    \caption{The system prompts for the planner.}
\end{small}
\end{table*}

\begin{table*}[h!]
\begin{small}
    \centering
    \begin{tabularx}{\textwidth}{|>{\centering\arraybackslash}X|}
    \hline
    \makecell[l]{    
    The information on the maps is: MAP INFORMATION \\
    The information of the units is UNIT INFORMATION \\
    All the units have no abilities, such as blinking or equipment.\\
    You win A times, tie B times, and lose C times out of D combats. \\
    There are E units and F enemy units left. \\
    You achieve G scores, give H damages to the enemy, take I damage on health, and take J damage on the shield. \\
    The history strategy and the results are: HISTORY \\
    To improve the winning rates, you may change a new tactic or add new tactics \\
    based on one of the historical strategies. You should provide me with at most 3 most important tactics.\\
    }\\
    \hline
    \end{tabularx}
    \captionsetup{labelformat=empty}
    \caption{The input prompt for the planner.}
\end{small}
\end{table*}

\clearpage
\subsection{Coder}

\begin{table*}[h!]
\begin{small}
    \centering
    \begin{tabularx}{\textwidth}{|>{\centering\arraybackslash}X|}
    \hline
    \makecell[l]{
    You are a StarCraft II player and a helpful assistant, and you are facing micro-management tasks.\\
I will describe the map and the units on the map.\\
You should not build any new buildings or new units or expand your base. \\
You should concentrate on the micro-management strategy.\\
I will give you the strategy in JSON array format. The keys are 'tactic\_name' and 'tactic\_description'.\\
You should implement the strategy in Python with burnysc2/pythonsc2 package.\\
You should concentrate on implementing the 'def async on\_step(self, iteration: int):' function.\\
The result should be surrounded in the ```python and ``` structure. \\
You should add some comments during the code generation for better explanation.\\
    }\\
    \hline
    \end{tabularx}
    \captionsetup{labelformat=empty}
    \caption{The system prompt for the coder.}
\end{small}
\end{table*}

\begin{table*}[h!]
\begin{small}
    \centering
    \begin{tabularx}{\textwidth}{|>{\centering\arraybackslash}X|}
    \hline
    \makecell[l]{
    TASK INFORMATION \\
    You should not use the await keyword.\\ 
    Make sure to check whether the list variables are empty or not. Do not use the run\_game function.\\
    The promotion from the critic module is PROMOTION \\
    The tactics from the planner module are: TACTICS \\
    Please implement the code for me. \\
    }\\
    \hline
    \end{tabularx}
    \captionsetup{labelformat=empty}
    \caption{The input prompt for the coder.}
\end{small}
\end{table*}

\subsection{Critic}

\begin{table}[h!]
\begin{small}
    \centering
    \begin{tabularx}{\textwidth}{|>{\centering\arraybackslash}X|}
    \hline
    \makecell[l]{
    You are a StarCraft II player and a helpful assistant, and you are facing micro-management tasks.\\
You are now working as a critic.\\
I will describe the map and the units on the map. You should not build any new buildings or new units. \\
After that, I will provide you with the tactic and the Python script which is the implementation of this tactic.\\
You should concentrate on the micro-management strategy to kill more enemies and preserve more units.\\
I will also provide you the result of the code, which might be the bug stack trace or the combat results.\\
You should analyze why the code leads to the result and \\
tell me the potential method to improve the performance based on the code. \\
You can suggest improving the current tactic or deleting some tactic based on the current code.\\
You do not need to provide me with the refinement code.\\
After that you should provide me 1 suggestion from ```[Change Tactic]``` or ```[Improve Tactic]```\\    
	}\\
    \hline
    \end{tabularx}
    \captionsetup{labelformat=empty}
    \caption{The system prompt for the critic.}
\end{small}
\end{table}

\begin{table*}[h!]
\begin{small}
    \centering
    \begin{tabularx}{\textwidth}{|>{\centering\arraybackslash}X|}
    \hline
    \makecell[l]{
    TASK INFORMATION \\
    The code is: \\
CODE \\
The result is: \\
RESULT \\
You should check whether the API you invoked follows the burnysc2/python-sc2 package.\\
Please summarize why the code caused the result.\\
Please briefly provide me the refinement of the tactic aiming at killing more enemy units \\
and cause more damage.\\
Do not show the revised code to me. \\
	}\\
    \hline
    \end{tabularx}
    \captionsetup{labelformat=empty}
    \caption{The input prompt for the critic.}
\end{small}
\end{table*}

\clearpage
\subsection{GRPO}

\begin{table}[h!]
\begin{small}
    \centering
    \begin{tabularx}{\textwidth}{|>{\centering\arraybackslash}X|}
    \hline
    \makecell[l]{
    You are a StarCraft II expert and AI agent. \\
    You will be given a micro-management scenario.\\
    Your task is to: \\
    1. Analyze the situation and propose a tactical plan\\
    2. Implement the plan as Python code using burnysc2/python-sc2 package\\
    Format your response EXACTLY as follows:\\
    <strategy> \\
    Write your tactical analysis and specific micro-management plan here\\
    </strategy>\\
    <code>\\
    ```python\\
    Write your implementation code here, focusing on the on\_step function\\
    ```\\
    </code>\\
    Important requirements: \\
    - Do not use await keywords\\
    - Always check if list variables are empty\\
    - Focus on micro-management only\\
    - No building or unit production \\    
	}\\
    \hline
    \end{tabularx}
    \captionsetup{labelformat=empty}
    \caption{The system prompt for the GRPO algorithm.}
\end{small}
\end{table}

\begin{table*}[h!]
\begin{small}
    \centering
    \begin{tabularx}{\textwidth}{|>{\centering\arraybackslash}X|}
    \hline
    \makecell[l]{
    TASK INFORMATION\\
    	}\\
    \hline
    \end{tabularx}
    \captionsetup{labelformat=empty}
    \caption{The input prompt for the critic.}
\end{small}
\end{table*}

\section{New SMAC Maps}

To validate the few-shot ability of the student LLM, we create new StarCraft II micromanagement maps. Each map is, to some extent, similar to some original SMAC tasks. Successfully generating correct scripts for these maps indicates that the student LLM is not overfitted and focuses on correct features. According to the figures below: 
\begin{enumerate}
	\item The 1mr1md\_vs\_1sc is similar to 2s\_vs\_1sc that agents should retreat to a safe point to recover health by Medivac instead of recovering shield automatically. 
	\item The 2vr\_vs\_2sc is also similar to the 2s\_vs\_1sc that the Void Rays should retreat to safe points to recover shield values.
	\item The 3mr\_3mr and the 5hd\_5hd are similar to the 3m and 8m scenario, in which the units are changed from Marine to Marauder and Hydralisk units.
	\item The 3rp\_v\_4zl is similar to the 2c\_vs\_64zg task that the Reaper units in 3rp\_vs\_4zl and the Colossus units in 2c\_vs\_64zg are able to jump (climb) up and down the cliff. The enemy Zealots and Zerglings can only reach by paths on the sides of the map.
	\item The 6mr\_vs\_1ac and the 6mr\_vs\_1st are similar to the so\_many\_baneling map. Agents should split to avoid splash damage from the enemies.
\end{enumerate}

\begin{figure}[h!]
    \centering
    \subfloat{\includegraphics[width=0.95\linewidth]{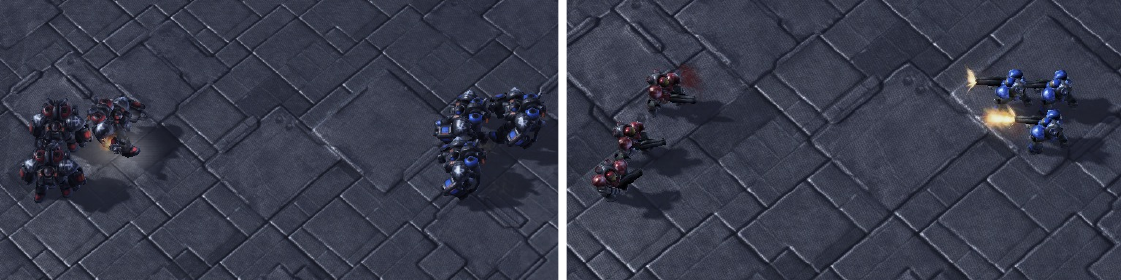}} 
    \newline
    \subfloat{\includegraphics[width=0.95\linewidth]{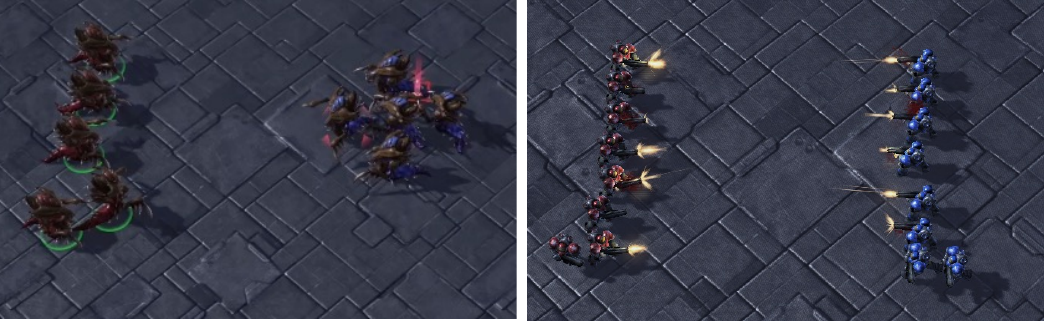}} 
    \newline
    \subfloat{\includegraphics[width=0.95\linewidth]{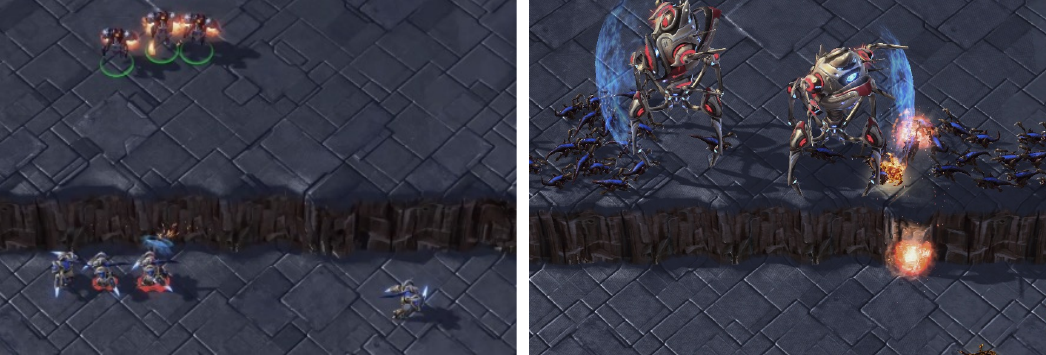}} 
    \newline
    \caption{The 7 newly designed SMAC maps to validate the student LLM. The left graphs are from the new maps, and the right graphs are from the original SMAC tasks. (1/2)}
    \label{fig:new_map} 
\end{figure}

\begin{figure}[h!]
    \centering
    \subfloat{\includegraphics[width=0.95\linewidth]{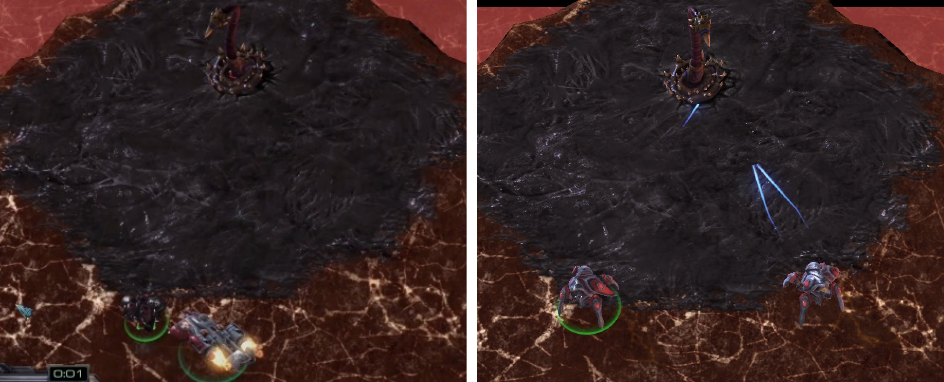}} 
    \newline
    \subfloat{\includegraphics[width=0.95\linewidth]{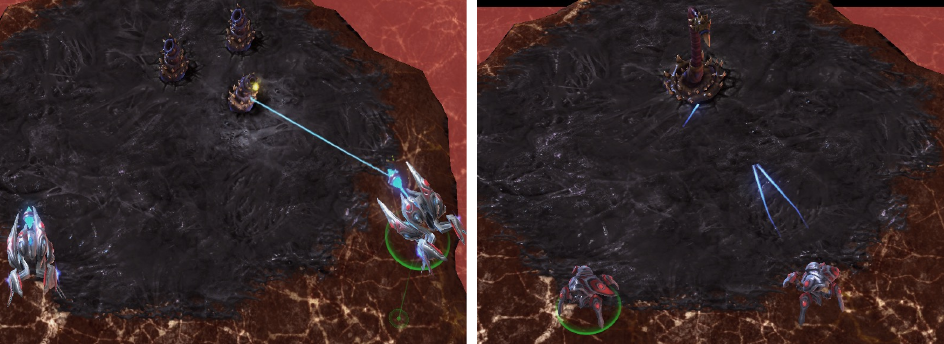}}
    \newline
    \subfloat{\includegraphics[width=0.95\linewidth]{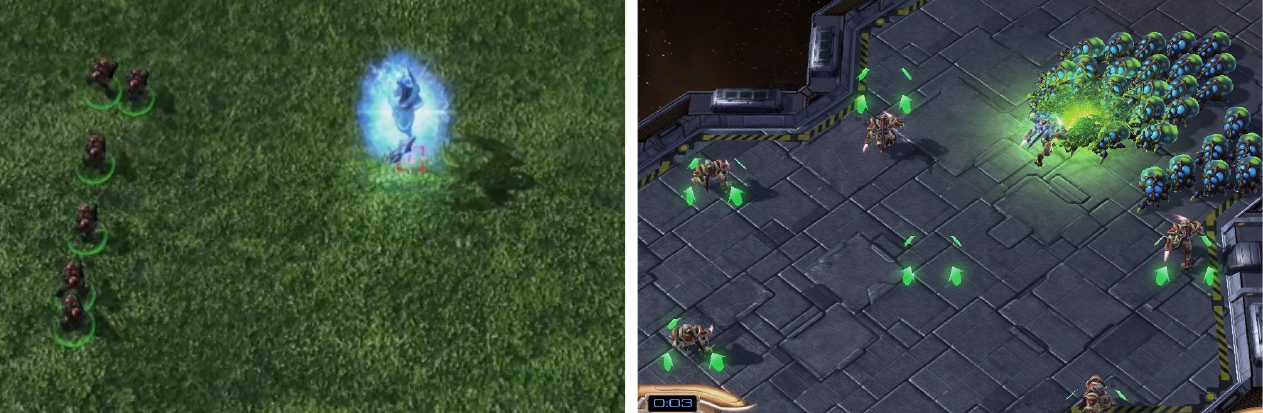}}
    \newline
    \subfloat{\includegraphics[width=0.95\linewidth]{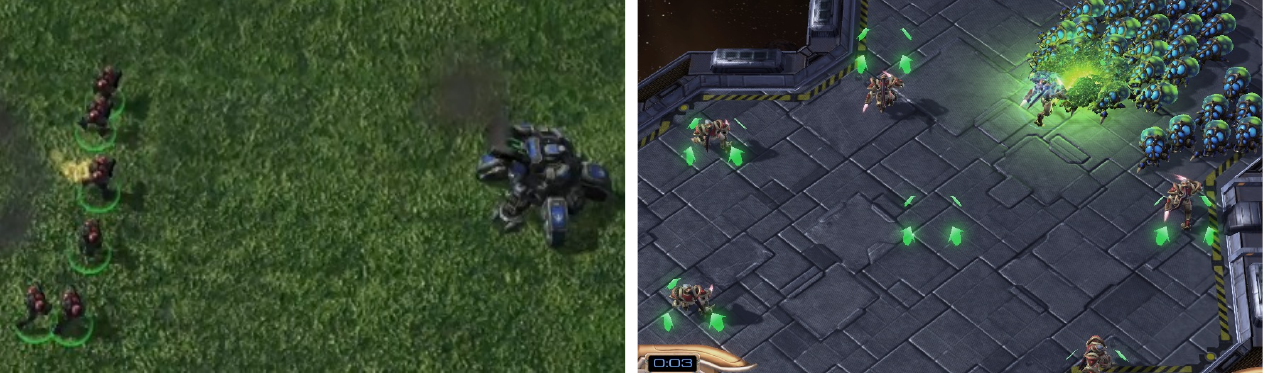}}
    \newline
    \caption{The 7 newly designed SMAC maps to validate the student LLM. The left graphs are from the new maps, and the right graphs are from the original SMAC tasks. (2/2)}
    \label{fig:new_map} 
\end{figure}

\clearpage
\section{Planner-Coder-Critic Case Study}

In this section, we take the 2s\_vs\_1sc map as an example to show the planner-coder-critic framework to solve SMAC tasks. The key strategy is the shield management skill that stalkers should retreat to safe positions to recover the shield value. 
\begin{table*}[h!]
    \centering
    \begin{small}
    \begin{tabularx}{\textwidth}{|>{\centering\arraybackslash}X|}
    \hline
    \makecell[l]{
 The map is 2s\_vs\_1sc.\\
You can control 2 Stalker units individually, and the enemy controls 1 Spine Crawler Structure.\\
The Stalker unit has 80 health, 80 shield, 1 defense, 6 attacking range, 4.13 speed, 13 damage with 9.7 DPS.\\
The Spine Crawler has 300 health, 0 shield, 2 defense, 7 attacking range, and 25 damage with 18.9 DPS.\\
All the units have no abilities, such as blinking or equipment.\\
The map is a 32*32-sized square map. \\
The available area of the x-axis is from 4 to 23, and the y-axis is from 7 to 30.\\
The enemy units are at (14, 21) points, and your Stalker units are at (11, 9) and (17, 9) points initially.\\
There are no terrain advantages or choke points in this map. You cannot get back to the enemy units.\\
	}\\
    \hline
    \end{tabularx}
    \end{small}
    \captionsetup{labelformat=empty}
    \caption{The task information in this showcase. We use the TASK INFORMATION placeholder in latter cases to prevent duplication.}
\end{table*}

\begin{table*}[h!]

    \centering
    \begin{small}
    \begin{tabularx}{\textwidth}{|>{\centering\arraybackslash}X|}
    \hline
    \makecell[l]{
You are a StarCraft II player and a helpful assistant, and you are facing micro-management tasks.\\
I will describe the map and the units on the map. You should not build any new buildings or new units. \\
You may focus on micro-management tactics to win the combat. \\
You should provide me the tactics in the format below:\\
\#\#\# Tactic 1: Tactic 1' name\\
**Condition to use:**\\
**Tactic Skeleton**\\
 \\
\#\#\# Tactic 2: Tactic 2' name\\
**Condition to use:**\\
**Tactic Skeleton**\\
 \\
Meanwhile, I will tell you history-taken tactics, and you may add new tactics onto the history tactics \\
by adding a new tactic to the list or try a new tactic. \\
 \\
TASK INFORMATION \\
\\
You should provide me with at most 3 most important tactics and describe the chosen tactic skeleton \\
in detail according to the situations of your unit and enemy units.\\
You should also indicate the condition to use this tactic. Make sure the conditions do not conflict.\\
	}\\
    \hline
    \end{tabularx}
    \end{small}
\end{table*}

We input the system prompt, the map information, and unit information into the planner. Then the planner outputs a strategy named 'Hit and Run'.

\begin{table*}[h!]
\begin{small}
    \centering
    \begin{tabularx}{\textwidth}{|>{\centering\arraybackslash}X|}
    \hline
    \makecell[l]{
    \#\#\# Tactic 1: Hit and Run\\
**Condition to use:** When the enemy units are within the Stalkers' attack range and the Stalkers \\ are not heavily damaged.\\
**Tactic Skeleton:**\\
1. Control both Stalkers to attack the Spine Crawler simultaneously.\\
2. After dealing damage, retreat slightly to avoid the Spine Crawler's counter-attack. \\
3. Repeat this process to slowly whittle down the Spine Crawler's health.\\
	}\\
    \hline
    \end{tabularx}
\end{small} 
\end{table*}

The planned strategy is then added to the prompt of the coder module.

\clearpage
\begin{table*}[h!]
\begin{small}
    \centering
    \begin{tabularx}{\textwidth}{|>{\centering\arraybackslash}X|}
    \hline
    \makecell[l]{
    You are a StarCraft II player and a helpful assistant, and you are facing micro-management tasks.\\
I will describe the map and the units on the map.\\
You should not build any new buildings or new units or expand your base. \\
You should concentrate on the micro-management strategy.\\
You should implement the strategy in Python with burnysc2/pythonsc2 package.\\
You should concentrate on implementing the 'def async on\_step(self, iteration: int):' function.\\
The result should be surrounded by the '```python' and '```' structure. \\
 \\
TASK INFORMATION\\
\\
You should not use the await keyword. Make sure to check whether the list variables are empty or not.\\
 \\
The tactic is: \\
\#\#\# Tactic 1: Hit and Run\\
**Condition to use:** When the enemy units are within the Stalkers' attack range and the Stalkers\\
are not heavily damaged.\\
**Tactic Skeleton:**\\
1. Control both Stalkers to attack the Spine Crawler simultaneously.\\
2. After dealing damage, retreat slightly to avoid the Spine Crawler's counter-attack. \\
3. Repeat this process to slowly whittle down the Spine Crawler's health.\\
\\
Please implement the code for me.\\
	}\\
    \hline
    \end{tabularx}
\end{small}
\end{table*}

Then, the coder takes the prompt with strategy as an input and outputs the generated code in the first round.
\begin{lstlisting}[language=Python]
from sc2.bot_ai import BotAI 
from sc2.ids.unit_typeid import UnitTypeId 
from sc2.position import Point2  
  
class HitAndRunBot(BotAI): 
	async def on_step(self, iteration: int):  
 		stalkers = self.units(UnitTypeId.STALKER)  
 		spine_crawler = self.enemy_structures.first 
 
 		if not spine_crawler or not stalkers.exists:  
 			return  
 
 		for stalker in stalkers: 
 			# Check if the Stalker is in attack range of the Spine Crawler  
 			if stalker.distance_to(spine_crawler) <= stalker.ground_range + 1:  
 				# Attack the Spine Crawler 
 				stalker.attack(spine_crawler)  
 			else:  
 				# Move towards the Spine Crawler if not in attack range 
 				stalker.move(spine_crawler.position) 
 			# After attacking, retreat slightly to avoid counter-attack  
 			if stalker.weapon_cooldown > 0: 
 				retreat_position = stalker.position.towards(self.start_location, 3) 
 				stalker.move(retreat_position)  
\end{lstlisting}

\clearpage
The generated scripts are stored as Python files and evaluated by the SMAC environment. The environment returns the result: You win 4 out of 10 times and lose 6 out of 10 times. 
You achieve 227.5 scores, give 280.0 damages to the enemy, take 140.6625 damage on health, and take 170.7625 damage on shield on average. After that, the results of the code are added to the prompt of the critic module to analyze the potential promotion.

\begin{table*}[h!]
\begin{small}
    \centering
    \begin{tabularx}{\textwidth}{|>{\centering\arraybackslash}X|}
    \hline
    \makecell[l]{
    You are a StarCraft II player and a helpful assistant, and you are facing micro-management tasks.\\
You are now working as a critic.\\
I will describe the map and the units on the map. You should not build any new buildings or new units. \\
After that, I will provide you with the tactic and the Python script which is the implementation of this tactic.\\
You should concentrate on the micro-management strategy to kill more enemies and preserve more units.\\
I will also provide you the result of the code, which might be the bug stack trace or the combat results.\\
You should analyze why the code leads to the result and tell me the potential method to improve \\
the performance based on the code. \\
You can suggest improving the current tactic or deleting some tactic based on the current code.\\
You do not need to provide me with the refinement code.\\
After that you should provide me a suggestion from ```[Change Tactic]``` or ```[Improve Tactic]```\\
 \\
TASK INFORMATION \\
\\
The code is: \\
CODE \\
 \\
The result is: \\
You Win 4 out of 10 times and lose 6 out of 10 times. You achieve 227.5 scores, give 280.0 damages\\
to the enemy, take 140.6625 damage on health, and take 170.7625 damage on shield on average.\\
 \\ 
You should check whether the API you invoked follows the burnysc2/python-sc2 package.\\
Please summarize why the code caused the result.\\
Meanwhile, please briefly provide me with 1 most important refinement of the tactic aiming at killing \\
more enemy units and cause more damage.\\
Do not show the revised code to me.\\
    }\\
    \hline
    \end{tabularx}
\end{small}
\end{table*}

\clearpage
The critic module analyzes the code and the results and outputs the potential refinement suggestion.

\begin{table*}[h!]
\begin{small}
    \centering
    \begin{tabularx}{\textwidth}{|>{\centering\arraybackslash}X|}
    \hline
    \makecell[l]{
\#\#\# Summary of the Result and Code Analysis  \\
 \\
The provided code implements a basic "hit-and-run" tactic where Stalkers attack the Spine Crawler and \\
then retreat slightly to avoid counter-attacks. However, the results show that this tactic is not very effective,\\
with a 40\% win rate and significant damage taken on both health and shields.\\
 \\
\#\#\#\# Why the Code Leads to the Result:\\
 \\
1. **Inconsistent Retreat Mechanism**:\\
 - The retreat mechanism is triggered only when the Stalker's weapon cooldown is greater than 0. \\
 This means that the Stalker will only retreat after attacking, but not necessarily after being attacked by the\\
 Spine Crawler. This can lead to situations where the Stalker takes multiple hits from the Spine Crawler\\
 before retreating, resulting in significant damage.\\
 \\
2. **No Consideration for Health and Shield Levels**:\\
 - The code does not consider the current health or shield levels of the Stalkers. If a Stalker's shield is \\
 depleted, it will take full damage from the Spine Crawler, which can quickly deplete its health.\\
 The tactic should prioritize retreating when shields are low or health is critically low.\\
 \\
3. **Fixed Retreat Distance**:\\
 - The retreat distance is fixed at 3 units towards the starting location. This might not be sufficient\\
  to ensure the Stalker is out of range of the Spine Crawler's attack, especially if the Spine Crawler\\
   is positioned close to the starting location.\\
 \\
4. **No Group Coordination**:\\
 - The code handles each Stalker individually without any coordination between them.\\
  This can lead to situations where one Stalker is retreating while the other is still attacking, \\
potentially exposing both to damage.\\
 \\
\#\#\#\# Potential Methods to Improve the Tactic: \\
 \\
1. **Dynamic Retreat Mechanism**:\\
 - Implement a more dynamic retreat mechanism that considers both the Stalker's weapon cooldown \\
 and its current health/shield levels. Retreat should be triggered not only after attacking\\
  but also when the Stalker's shields are low or health is critically low.\\
 \\
2. **Variable Retreat Distance**:\\
 - Use a variable retreat distance based on the Stalker's current position relative to the Spine Crawler.\\
  Ensure that the retreat distance is sufficient to get the Stalker out of the Spine Crawler's attack range.\\
 \\
3. **Group Coordination**:\\
 - Coordinate the actions of both Stalkers to ensure that they are not both attacking or retreating at the same\\
 time. This can help in managing the Spine Crawler's attention and reducing overall damage taken.\\
 \\
4. **Shield Management**:\\
 - Prioritize retreating when shields are low to prevent the Stalker from taking full damage from the \\
 Spine Crawler. Consider using a shield recharge mechanic if available in the environment.
 \\
\#\#\# Suggestion:\\
```[Improve Tactic]```\\
    }\\
    \hline
    \end{tabularx}
\end{small}
\end{table*}

\clearpage
Because the [Improve Tactic] is given instead of [Change Tactic], the coder takes the critic output as the prompt and re-generates the code.

\begin{table*}[h!]
\begin{small}
    \centering
    \begin{tabularx}{\textwidth}{|>{\centering\arraybackslash}X|}
    \hline
    \makecell[l]{
    TASK INFORMATION\\
    \\
    You should not use the await keyword.\\
    Make sure to check whether the list variables are empty or not.\\
    \\
 	CRITIC SUGGESTION \\
 	\\
 	Please implement the code for me.\\
	}\\
    \hline
    \end{tabularx}
\end{small}
\end{table*}

Then, the coder generates new codes according to the refinement suggestion given by the critic module.

\begin{lstlisting}[language=Python]
from sc2.bot_ai import BotAI
from sc2.ids.unit_typeid import UnitTypeId
from sc2.position import Point2
 
class HitAndRunBot(BotAI):
	async def on_step(self, iteration: int):
 		stalkers = self.units(UnitTypeId.STALKER)
 		spine_crawler = self.enemy_structures.first
 
 		if not spine_crawler or not stalkers.exists:
 			return
 
		for stalker in stalkers:
 			# Check if the Stalker is in attack range of the Spine Crawler
 			if stalker.distance_to(spine_crawler) <= stalker.ground_range + 1:
 				# Attack the Spine Crawler
  				stalker.attack(spine_crawler)
 			else:
 				# Move towards the Spine Crawler if not in attack range
 				stalker.move(spine_crawler.position)
 
 			# After attacking, retreat slightly to avoid counter-attack
 			if stalker.weapon_cooldown > 0 or stalker.shield_percentage < 0.2 or stalker.health_percentage < 0.2:
 				retreat_distance = 5 # Adjust retreat distance based on current position
 				retreat_position = stalker.position.towards(self.start_location, retreat_distance)
 				stalker.move(retreat_position)
 
 			# Group coordination: Ensure both Stalkers are not both attacking or retreating at the same time
  			if len(stalkers) == 2:
 				stalker1, stalker2 = stalkers
 				if stalker1.weapon_cooldown > 0 and stalker2.weapon_cooldown == 0:
 					stalker2.attack(spine_crawler)
 				elif stalker2.weapon_cooldown > 0 and stalker1.weapon_cooldown == 0:
 					stalker1.attack(spine_crawler)
\end{lstlisting}

Finally, the newly generated code script is stored as a Python file and tested through the SMAC environment. The environment outputs the winning result: You win 10 out of 10 times and lose 0 out of 10 times. 
You achieve 350.0 scores, give 331.2 damages to the enemy, take 115.3125 damage on health, and take 405.0875 damage on shield on average. The winning rate achieves 100\% and the pipeline terminates.

\clearpage
\section{GRPO Case Study (2s3z Map)}

\subsection{Sample 1: Score:-0.5}

\begin{table*}[h!]
\begin{small}
    \centering
    \begin{tabularx}{\textwidth}{|>{\centering\arraybackslash}X|}
    \hline
    \makecell[l]{
    <strategy>\\
The scenario presents a balanced early game with both players having similar units. \\
The key to success lies in effectively micro-managing the units to harass, disrupt, and gain an advantage. \\
Here's the plan:\\
\\
1. Positioning: Place Stalkers slightly behind the Zealots to provide cover and allow for better positioning.\\
2. Early harassment: Use Stalkers to harass the enemy units, focusing on weakening them before engaging.\\
3. Zealot engagement: Use Zealots to engage the enemy units, taking advantage of their high health\\
and damage output.\\
4. Split Stalker harassment: Split the Stalker units to harass the enemy units from different directions,\\
 causing more damage and forcing the enemy to split their units to defend.\\
5. Zealot support: Use Zealots to support Stalker harassment by engaging enemy units while Stalkers focus \\
on weakening them.\\
6. Retreat and reposition: If the enemy units are too strong, retreat and reposition to avoid \\
unnecessary losses.\\
</strategy>\\
\\
Now, let's implement this plan in the on\_step function:\\
	}\\
    \hline
    \end{tabularx}
\end{small}
\end{table*}

\begin{lstlisting}[language=Python]
from sc2 import maps
from sc2.bot_ai import BotAI
from sc2.data import Race, Difficulty
from sc2.ids.ability_id import AbilityId
from sc2.ids.effect_id import EffectId
from sc2.ids.unit_typeid import UnitTypeId
from sc2.main import run_game
from sc2.player import Bot, Computer
from sc2.position import Point2
from sc2.unit import Unit
from sc2.units import Units
from math import cos, sin, pi
import math
import random

class BattleBot(BotAI):

    def __init__(self):
        super().__init__()
        self.stalker_positions = []
        self.zealot_positions = []
        self.target_position = Point2((16, 16))  # Center of the map

    async def on_step(self, step):
        # Get units
        my_stalkers = self.units(UnitTypeId.STALKER)
        my_zealots = self.units(UnitTypeId.ZEALOT)
        enemy_units = self.enemy_units

        # Update positions
        self.stalker_positions = [stalker.position for stalker in my_stalkers]
        self.zealot_positions = [zealot.position for zealot in my_zealots]

        # Early harassment with Stalkers
        if step < 100:
            for stalker in my_stalkers:
                closest_enemy = enemy_units.closest_to(stalker)
                if closest_enemy and stalker.distance_to(closest_enemy) > stalker.ground_range:
                    stalker.attack(closest_enemy)
                else:
                    stalker.move(self.target_position)

        # Zealot engagement and support
        for zealot in my_zealots:
            closest_enemy = enemy_units.closest_to(zealot)
            if closest_enemy and zealot.distance_to(closest_enemy) <= zealot.ground_range:
                zealot.attack(closest_enemy)

        # Split Stalker harassment
        if len(enemy_units) > 0:
            for stalker in my_stalkers:
                closest_enemy = enemy_units.closest_to(stalker)
                if closest_enemy and stalker.distance_to(closest_enemy) > stalker.ground_range:
                    stalker.attack(closest_enemy)
                else:
                    stalker.move(self.target_position)

        # Retreat and reposition
        for unit in self.units:
            if unit.health_percentage < 0.5:
                unit.move(self.target_position)

\end{lstlisting}

\subsection{Sample 2: Score:0.0}

\begin{table*}[h!]
\begin{small}
    \centering
    \begin{tabularx}{\textwidth}{|>{\centering\arraybackslash}X|}
    \hline
    \makecell[l]{
    <strategy>\\
The key to success in this scenario is to effectively harass and disrupt the enemy units while maintaining \\
a strong defensive posture. \\
We will use a combination of Stalker and Zealot micro-management to achieve this goal.\\
</strategy>\\
	}\\
    \hline
    \end{tabularx}
\end{small}
\end{table*}

\begin{lstlisting}[language=Python]
from sc2 import maps
from sc2.bot_ai import BotAI
from sc2.data import Race, Difficulty
from sc2.ids.ability_id import AbilityId
from sc2.ids.effect_id import EffectId
from sc2.ids.unit_typeid import UnitTypeId
from sc2.main import run_game
from sc2.player import Bot, Computer
from sc2.position import Point2
from sc2.unit import Unit
from sc2.units import Units
from math import cos, sin, pi
import math
import random

class BattleBot(BotAI):

    async def on_step(self, step):
        # Get all units
        stalkers = self.units(UnitTypeId.STALKER).idle
        zealots = self.units(UnitTypeId.ZEALOT).idle
        enemy_units = self.enemy_units

        # Stalker Micro-Management
        for stalker in stalkers:
            closest_enemy = enemy_units.closest_to(stalker)
            if closest_enemy:
                if stalker.distance_to(closest_enemy) > stalker.ground_range:
                    stalker.move(closest_enemy)
                elif stalker.health_percentage < 0.5:
                    stalker.move(Point2((9.0, 16.0)))
                else:
                    stalker.attack(closest_enemy)

        # Zealot Micro-Management
        for zealot in zealots:
            closest_enemy = enemy_units.closest_to(zealot)
            if closest_enemy:
                if zealot.distance_to(closest_enemy) > 0.1:
                    zealot.move(closest_enemy)
                elif zealot.health_percentage < 0.5:
                    zealot.move(Point2((9.0, 16.0)))
                else:
                    zealot.attack(closest_enemy)

        # Additional logic to handle enemy units
        if enemy_units:
            for unit in self.units:
                if unit.is_idle:
                    closest_enemy = enemy_units.closest_to(unit)
                    if closest_enemy:
                        unit.attack(closest_enemy)
   
\end{lstlisting}

\subsection{Sample 3: Score:0.2}

\begin{table*}[h!]
\begin{small}
    \centering
    \begin{tabularx}{\textwidth}{|>{\centering\arraybackslash}X|}
    \hline
    \makecell[l]{
    <strategy>\\
The key to success in this scenario is to create a dynamic and responsive defensive line using \\
the Stalker and Zealot units. \\
We will focus on harassing the enemy units, retreating when necessary, and repositioning to counter \\
the enemy's movements. \\
By maintaining strict control over our units' movements and attack patterns, we can force the enemy\\
to waste resources and time, potentially leading to a stalemate or even an advantage.\\
</strategy>\\
	}\\
    \hline
    \end{tabularx}
\end{small}
\end{table*}

\begin{lstlisting}[language=Python]
from sc2 import maps
from sc2.bot_ai import BotAI
from sc2.data import Race, Difficulty
from sc2.ids.ability_id import AbilityId
from sc2.ids.effect_id import EffectId
from sc2.ids.unit_typeid import UnitTypeId
from sc2.main import run_game
from sc2.player import Bot, Computer
from sc2.position import Point2
from sc2.unit import Unit
from sc2.units import Units
from math import cos, sin, pi
import math
import random

class BattleBot(BotAI):

    async def on_step(self, step):
        # Collect our units
        stalkers = self.units(UnitTypeId.STALKER).idle
        zealots = self.units(UnitTypeId.ZEALOT).idle

        # Collect enemy units
        enemy_units = self.enemy_units
        enemy_stalkers = enemy_units.of_type(UnitTypeId.STALKER)
        enemy_zealots = enemy_units.of_type(UnitTypeId.ZEALOT)

        # Define the initial positions
        self.center_position = Point2((16.0, 16.0))
        self.front_line = Point2((16.0, 18.0))
        self.rear_position = Point2((16.0, 14.0))

        # Stalker Micro Management
        if not stalkers.empty:
            # Prioritize harassing the closest enemy stalker
            closest_enemy_stalker = enemy_stalkers.closest_to(self.center_position)
            for stalker in stalkers:
                if stalker.distance_to(closest_enemy_stalker) < 5:
                    stalker.attack(closest_enemy_stalker)
                else:
                    stalker.move(self.front_line)
            # Move Stalkers to the front line if they are not engaging
            for stalker in stalkers:
                if stalker.distance_to(closest_enemy_stalker) >= 5:
                    stalker.move(self.front_line)

        # Zealot Micro Management
        if not zealots.empty:
            # Prioritize engaging all enemy zealots simultaneously
            for zealot in zealots:
                if not enemy_zealots.empty:
                    closest_enemy_zealot = enemy_zealots.closest_to(zealot)
                    zealot.attack(closest_enemy_zealot)
                    # Move Zealots to the front line if they are not engaging
                    if zealot.distance_to(closest_enemy_zealot) >= 5:
                        zealot.move(self.front_line)
   
\end{lstlisting}

\subsection{Sample 4: Score:1.0}

\begin{table*}[h!]
\begin{small}
    \centering
    \begin{tabularx}{\textwidth}{|>{\centering\arraybackslash}X|}
    \hline
    \makecell[l]{
    <strategy>\\
The key to success in this scenario is to create a dynamic and responsive defensive line using the \\
Stalker and Zealot units.\\
 We will focus on harassing the enemy units, retreating when necessary, and repositioning to counter \\
 the enemy's movements.\\
 By maintaining strict control over our units' movements and attack patterns, we can force the enemy\\
 to waste resources and time, potentially leading to a stalemate or even an advantage.\\
</strategy>\\
	}\\
    \hline
    \end{tabularx}
\end{small}
\end{table*}

\begin{lstlisting}[language=Python]
from sc2 import maps
from sc2.bot_ai import BotAI
from sc2.data import Race, Difficulty
from sc2.ids.ability_id import AbilityId
from sc2.ids.effect_id import EffectId
from sc2.ids.unit_typeid import UnitTypeId
from sc2.main import run_game
from sc2.player import Bot, Computer
from sc2.position import Point2
from sc2.unit import Unit
from sc2.units import Units
from math import cos, sin, pi
import math
import random

class BattleBot(BotAI):

    async def on_step(self, step):
        # Collect our units
        stalkers = self.units(UnitTypeId.STALKER).idle
        zealots = self.units(UnitTypeId.ZEALOT).idle

        # Collect enemy units
        enemy_units = self.enemy_units
        enemy_stalkers = enemy_units.of_type(UnitTypeId.STALKER)
        enemy_zealots = enemy_units.of_type(UnitTypeId.ZEALOT)

        # Define the initial positions
        self.center_position = Point2((16.0, 16.0))
        self.front_line = Point2((16.0, 18.0))
        self.rear_position = Point2((16.0, 14.0))

        # Stalker Micro Management
        if not stalkers.empty:
            # Prioritize harassing the closest enemy stalker
            closest_enemy_stalker = enemy_stalkers.closest_to(self.center_position)
            for stalker in stalkers:
                if stalker.distance_to(closest_enemy_stalker) < 5:
                    stalker.attack(closest_enemy_stalker)
                else:
                    stalker.move(self.front_line)
            # Move Stalkers to the front line if they are not engaging
            for stalker in stalkers:
                if stalker.distance_to(closest_enemy_stalker) >= 5:
                    stalker.move(self.front_line)

        # Zealot Micro Management
        if not zealots.empty:
            # Prioritize engaging all enemy zealots simultaneously
            for zealot in zealots:
                if not enemy_zealots.empty:
                    closest_enemy_zealot = enemy_zealots.closest_to(zealot)
                    zealot.attack(closest_enemy_zealot)
                    # Move Zealots to the front line if they are not engaging
                    if zealot.distance_to(closest_enemy_zealot) >= 5:
                        zealot.move(self.front_line)
   
\end{lstlisting}

\end{document}